# WHU-STree: A Multi-modal Benchmark Dataset for Street Tree Inventory

Ruifei Ding[a,1], Zhe Chen[a,1], Wen Fan[b], Chen Long[a], Huijuan Xiao[c], Yelu Zeng[d,e], Zhen Dong[a,*], Bisheng Yang[a]

[a]*State Key Laboratory of Information Engineering in Surveying, Mapping and Remote Sensing, Wuhan University, Wuhan, 430079, China*
[b]*Institute of Forest Management, TUM School of Life Sciences Weihenstephan, Technical University of Munich, Freising, 85354, Germany*
[c]*Department of Civil and Environmental Engineering, The Hong Kong University of Science and Technology, Hong Kong SAR, China*
[d]*College of Land Science and Technology, China Agricultural University, Beijing, 100083, China*
[e]*Key Laboratory of Remote Sensing for Agri-Hazards, Ministry of Agriculture and Rural Affairs, Beijing, 100083, China*

**Abstract**

Street trees are vital to urban livability, providing ecological and social benefits. Establishing a detailed, accurate, and dynamically updated street tree inventory has become essential for optimizing these multifunctional assets within space-constrained urban environments. Given that traditional field surveys are time-consuming and labor-intensive, automated surveys utilizing Mobile Mapping Systems (MMS) offer a more efficient solution. However, existing MMS-acquired tree datasets are limited by small-scale scene, limited annotation, or single modality, restricting their utility for comprehensive analysis. To address these limitations, we introduce WHU-STree, a cross-city, richly annotated, and multi-modal urban street tree dataset. Collected across two distinct cities, WHU-STree integrates synchronized point clouds and high-resolution images, encompassing 21,007 annotated tree instances across 50 species and 2 morphological parameters. Leveraging the unique characteristics, WHU-STree concurrently supports over 10 tasks related to street tree inventory. We benchmark representative baselines for two key

---

*Corresponding author: dongzhenwhu@whu.edu.cn
[1]The two authors contribute equaly to this work.

tasks—tree species classification and individual tree segmentation. Extensive experiments and in-depth analysis demonstrate the significant potential of multi-modal data fusion and underscore cross-domain applicability as a critical prerequisite for practical algorithm deployment. In particular, we identify key challenges and outline potential future works for fully exploiting WHU-STree, encompassing multi-modal fusion, multi-task collaboration, cross-domain generalization, spatial pattern learning, and Multi-modal Large Language Model for street tree asset management. The WHU-STree dataset is accessible at: https://github.com/WHU-USI3DV/WHU-STree.

## 1. Introduction

Street trees, vital to urban ecosystems, provide ecological benefits (e.g., shade (Kumar et al., 2024), air purification (Grundström and Pleijel, 2014), noise reduction (Salmond et al., 2016)) and social advantages (e.g., public health promotion (Wolf et al., 2020; Pataki et al., 2021), crime reduction (Troy et al., 2012), road safety enhancement (Kondo et al., 2017)). These benefits are closely tied to the stock, structure, and species of the trees (Grote et al., 2016; Liu et al., 2023b). However, the increasing scarcity of urban land competes with the demand for street trees and their ecosystem services (Seiferling et al., 2017). Maximizing service potential under spatial constraints necessitates a comprehensive, dynamically updated street tree inventory. The street tree inventory records diverse attributes, including location, species, growth conditions, morphological parameters, etc (Nielsen et al., 2014; Ma et al., 2021). Traditionally, these attributes are collected through field surveys by trained staff (Berland and Lange, 2017). Although field surveys yield highly accurate data, their time-consuming and labor-intensive characteristics limit the ability to meet the growing demand for large-scale, rapidly updated street tree inventories (Alonzo et al., 2016).

In urban areas, Mobile Mapping Systems (MMS) equipped with laser scanners and cameras possess close-range perception capabilities (Yang et al., 2024b). These systems can simultaneously acquire point cloud and image data, thereby facilitating the inventory of street assets (Nassar et al., 2020; Fang et al., 2022; Zhou et al., 2022). However, existing urban scene datasets acquired by MMS pay insufficient attention to street trees, hindering the development of automated inventory solutions based on Deep Learning (DL). Trees are typically annotated as a single semantic category without instance-level labels, and species information is often omitted (Roynard et al., 2018; Behley et al., 2019; Tan et al., 2020; Han et al., 2024). Although some datasets can be utilized for tasks such as individual tree segmentation (Fan et al., 2021), species classification (Choi et al., 2022; Liu et al., 2023b), and morphological parameter estimation (Wu et al., 2013; Wang et al., 2018), their respective limitations prevent them from serving as benchmarks for multi-task joint processing, resulting in a fragmented investigation of urban street tree assets. To the best of our knowledge, there are currently only two datasets that support multi-task processing, specifically for individual street tree species classification and individual tree segmentation: Tree-ML (Yazdi et al., 2024) and TDUS (Yang et al., 2023). However, their uni-modal nature limits their effectiveness for tasks requiring multi-modal data.

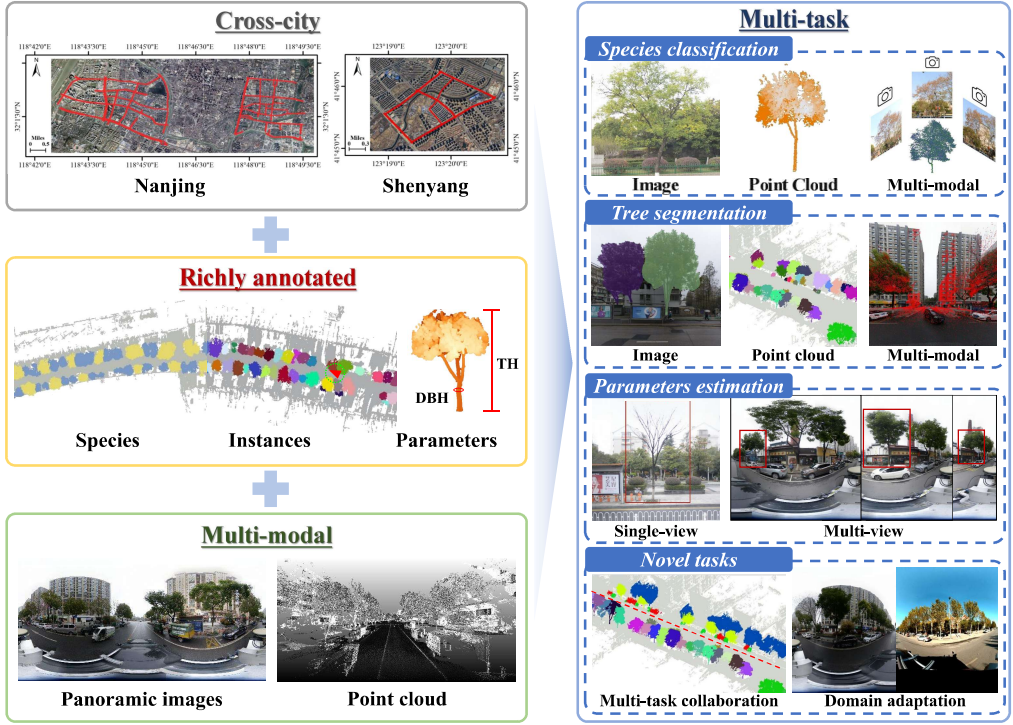

Fig. 1. Overview of the WHU-STree dataset. The cross-city, richly annotated, and multi-modal dataset concurrently supports over 10 tasks.

Tree-ML, relying on point clouds, lacks image-based texture and spectral information, hindering fine-grained species classification (Yazdi et al., 2024). Conversely, TDUS, an image-based dataset, lacks point cloud data, limiting accurate tree instance localization and morphological parameter estimation (Yang et al., 2023). Furthermore, these datasets, typically collected from a single region, restrict cross-domain studies and hinder validation of algorithm generalization and robustness.

In summary, current urban street tree datasets exhibit the following limitations. (1) **Small-scale scene**: data splits within a single scene limit the assessment of algorithm generalization, critical for real-world applications. (2) **Limited annotation**: street trees encompass diverse attributes, and datasets with limited annotations restrict their applicability for comprehensive asset inventories. (3) **Single modality**: existing datasets are typically based on a single modality, constraining their performance across various tasks. There is a need to explore the benefits of multi-modal information.

To overcome the aforementioned limitations, we present a multi-modal, richly-annotated, and cross-city urban street tree dataset, WHU-STree. Leveraging these features, the dataset supports multiple tasks simultaneously, enabling comprehensive street tree inventories and facilitating the development of novel task settings, such as estimating 3D tree morphology from 2D images. Additionally, the dataset provides both point cloud and image data, allowing for complementary advantages between the two modalities, which can facilitate the exploration of the relationships between multi-modal data. Furthermore, the cross-city data enables the evaluation of algorithm robustness and the development of methods with strong generalization capabilities. Specifically, the main contributions of this study are summarized as follows:

(1) We present a multi-modal urban street tree dataset, WHU-STree, which integrates both point cloud and image data across two distinct cities. It contains 21,007 annotated tree instances across 50 species and 2 morphological parameters. Leveraging its comprehensive strengths, the dataset supports a wide range of tasks related to street tree inventories.
(2) We provide detailed baselines for two tasks (i.e., tree species classification and individual tree segmentation), thoroughly investigating the performance and generalization ability of existing algorithms, and highlighting the importance of multi-modal fusion and multi-task collaboration.
(3) We analyze the current challenges and opportunities from the perspectives of multi-modal fusion, multi-task collaboration, cross-domain gen-eralization, spatial pattern learning, and Multi-modal Large Language Model (MLLM) for street tree asset management. We hope this dataset can serve as a benchmark to promote further research on automated street tree inventories.

## 2. Related Works

Given the significance of tree asset inventory, numerous specialized algorithms for tree species classification and individual tree segmentation have been developed. In line with this trend, several datasets have been introduced to standardize evaluation and promote progress in this field.

### 2.1. Existing datasets

Various data modalities, such as point clouds and images, have been utilized for tree species classification and individual tree segmentation. Table 1 summarizes the mainstream datasets and their key characteristics.

#### 2.1.1. Datasets for tree species classification

Algorithms for tree species classification using point clouds are predominantly validated on private datasets in existing literature, which are rarely made publicly available, impeding comparative analysis and hindering progress in the field. For-species20k dataset (Puliti et al., 2024) addresses this gap by providing the first benchmark for point cloud-based tree species classification. It includes over 20,000 trees from 33 species across various forest types and features diverse tree sizes and shapes. While point clouds offer rich geometric information for capturing tree morphology, the spectral and texture characteristics in images are often more effective for species classification. Numerous studies have utilized close-up images of leaves and bark for this purpose (Kumar et al., 2012; Hervé Goëau, 2012; Carpentier et al., 2018; Ratajczak et al., 2019). Although these approaches achieve high accuracy, the reliance on close-range photography limits their suitability for large-scale and automated inventories. In addition, several datasets leverage multi-source data. Wegner et al. (2016) collected "Pasadena Urban Trees" dataset, which includes geographic and species annotations for over 80,000 trees, along with aerial and street-view imagery. TreeSatAI (Ahlswede et al., 2022) provides a benchmark for 20 tree species classification using multi-sensor data from aerial, Sentinel-1, and Sentinel-2 imagery. However, instance-level annotations are not available in these datasets.

Table 1
Exisiting typical dataset for tree inventory.

| Dataset | Scene | Quantities | Data Type | | Cross-region | Annotation | | |
|---|---|---|---|---|---|---|---|---|
| | | | *Point cloud* | *Image* | | *Instances* | *Species* | *Parameters* |
| Pasadena Urban Trees (Wegner et al., 2016) | Urban | 80K | - | RGB | | ✓ | 18 | |
| Wytham woods (Calders et al., 2022) | Forest | 835 | TLS | - | | ✓ | - | ✓ |
| LAUTx (Andreas et al., 2022) | Forest | 515 | MLS | - | | ✓ | - | ✓ |
| SYSSIFOSS (Weiser et al., 2022) | Forest | 1491 | ALS& ULS& TLS | - | ✓ | ✓ | 22 | ✓ |
| TreeSatAI Ahlswede et al. (2022) | Forest | - | - | RGB & SAR & Multispctral | ✓ | | 20 | |
| NeonTreeEvaluation (Ben Weinstein and White, 2022) | Forest | 31K+ | ALS | RGB & Hyperspectral | ✓ | ✓ | - | |
| TDUS (Yang et al., 2023) | Urban | - | - | RGB | ✓ | ✓ | 50 | |
| Treelearn (Henrich et al., 2023) | Forest | 235 | MLS | - | | ✓ | - | |
| FOR-instance (Puliti et al., 2023) | Forest | 1130 | ULS | - | ✓ | ✓ | - | ✓ |
| MillionTrees (Weinstein, 2023) | Forest & Urban | - | - | RGB | ✓ | ✓ | 21+ | |
| ForestSemantic (Liang et al., 2024) | Forest | 673 | TLS | - | | ✓ | 3 | |
| OAM-TCD (Veitch-Michaelis et al., 2024) | Forest & Urban | 280K+ | - | RGB | ✓ | ✓ | - | |
| RT-Trees (Kapil et al., 2024) | Forest | - | - | RGB-thermal | | ✓ | - | |
| Tree-ML (Yazdi et al., 2024) | Urban | 3755 | MLS | - | | ✓ | 20+ | ✓ |
| FOR-spices20k (Puliti et al., 2024) | Forest | 20158 | TLS & MLS & ULS | - | ✓ | | 33 | |
| WHU-STree (Ours) | Urban | 21007 | MLS | RGB | ✓ | ✓ | 50 | ✓ |

*2.1.2. Datasets for tree segmentation*

Due to the ability to directly depict the 3D information, point clouds are often used for tree segmentation and morphological parameter estimation, which facilitates biomass estimation. Existing research on tree segmentation predominantly focuses on forest scenes, where trees are densely distributed and the surroundings are relatively homogeneous. Consequently, datasets for forest scenes emerged earlier and are more numerous (Calders et al., 2022; Andreas et al., 2022). SYSSIFOSS (Weiser et al., 2022) comprises 1,491 trees with species information, collected using three different methods (ALS, ULS, TLS). Treelearn (Henrich et al., 2023) contains data collected from MLS in German forests, with 235 manually labeled trees. However, its limited size restricts its applicability for DL-based methods. FOR-instance (Puliti et al., 2023) and ForestSemantic (Liang et al., 2024) include ULS and TLS point clouds, respectively, and provide fine-grained component-level and instance-level annotations, but lack species information. Additionally, aerial imagery has also been used for individual tree detection. MillionsTrees (Weinstein, 2023) aggregates open accessed datasets to create a unified benchmark for airborne tree detection. RT-Trees (Kapil et al., 2024) includes RGB and thermal infrared (TIR) imagery using drones in mixed forest regions of central Canada. OAM-TCD (Veitch-Michaelis et al., 2024) presents a global dataset based on VHR images, annotating over 280,000 individual trees and 56,000 tree clusters, covering various urban and natural environments. In addtion, NeonTreeEvaluation (Ben Weinstein and White, 2022), built upon the National Ecological Observation Network (NEON) (Ben Weinstein and White, 2024), provides LiDAR, aerial RGB, and hyperspectral imagery with bounding-box annotations. However, due to the significant differences in scene complexity, these datasets are primarily suitable as auxiliary resources for urban street tree research. Despite the relatively few studies on street trees, interest in urban scenes remains strong. Notable datasets have emerged to date, such as SemanticKitti (Behley et al., 2019) and Toronto-3D (Tan et al., 2020). However, in these datasets, trees are often labeled as a single semantic class. Paris-Lille-3D (Roynard et al., 2018) and WHU-Urban3D (Han et al., 2024) further provide instance-level annotations for trees but do not provide species information. Similarly, images can also be used for urban street tree understanding, such as Cityscapes (Cordts et al., 2016) and ADE20K (Zhou et al., 2019). However, these datasets do not distinguish individual instances and lack species information.

To the best of our knowledge, only two datasets currently support multi-task urban street tree inventory, encompassing both tree species classification and individual tree segmentation. TreeML-Data (Yazdi et al., 2024) comprises MLS data collected from 40 streets in Munich, including 3,755 tree instances with species labels, and constructs a graph-based model for each tree. However, it lacks co-registered imaging data, which limits its effectiveness for tree species classification. Yang et al. (2023) constructed an urban street tree dataset containing both detailed (leaf, bark, branch) and holistic information of trees, supporting tasks such as species classification and instance segmentation. Nevertheless, the absence of point cloud hinders the direct extraction of morphological parameters from the dataset.

## 2.2. Deep learning-based methods

Many previous studies have applied machine learning to tree species classification and individual tree segmentation on images (Jing et al., 2012; Huang et al., 2018; Persson et al., 2018). Incorporating 3D features extracted from point cloud data has also been shown to improve model performance (Liu et al., 2017; Liang et al., 2018; Hartling et al., 2021; Qin et al., 2022). However, due to the complexity of manual feature extraction and the difficulty in assessing feature importance, machine learning has gradually been supplanted by DL (Chen et al., 2024a).

### 2.2.1. Single-modality methods

At present, U-Net-based models for semantic segmentation (Ronneberger et al., 2015; Schiefer et al., 2020; Sun et al., 2022; Li et al., 2023b) and RCNN-based models (He et al., 2017; Ocer et al., 2020; Xi et al., 2021) for instance segmentation have been widely applied to image-based tree segmentation. For instance, Schiefer et al. (2020) proposed a CNN-based method for mapping forest tree species using UAV-acquired RGB images. Freudenberg et al. (2019) utilized a U-Net architecture for large-scale detection of palm trees in VHR images. Ocer et al. (2020) employed a Mask R-CNN model combined with a Feature Pyramid Network (FPN) for individual tree segmentation from UAV-acquired RGB images. With recent advancements, researchers have used transformer-based models (Dosovitskiy, 2020; Amirkolaee et al., 2023) and vision foundation models like Segment Anything (SAM) (Osco et al., 2023; Kirillov et al., 2023) for tree segmentation, demonstrating strong potential.

Due to the relatively slower progress in 3D segmentation algorithms, early DL-based studies projected the original point clouds onto a 2D plane using top-down (Windrim and Bryson, 2019; Chang et al., 2022) or multi-view projections (Seidel et al., 2021; Briechle et al., 2021) for processing, and subsequently re-projected the results back into 3D space. However, these projection-based algorithms compromise the representation of spatial features. To fully exploit spatial information, many DL-based methods have been developed that directly process raw point clouds for tree segmentation or species classification. Some studies (Liu et al., 2021, 2022a), inspired by the PointNet series (Qi et al., 2017a,b), classify tree species by directly feeding individual tree point clouds into point-based networks. Additionally, voxel-based and attention-based methods have also been explored and applied to tree species classification tasks (Chen et al., 2021; Liu et al., 2022b).

For individual tree segmentation, some studies predict the offset of points to tree centers (Luo et al., 2021) or centroids (Jiang et al., 2023) using networks, followed by clustering. Nevertheless, these methods often require explicit tree point extraction and multiple processing stages, which may lead to error accumulation throughout the pipeline. To mitigate this issue, end-to-end instance segmentation networks (Vu et al., 2022; Sun et al., 2023; Schult et al., 2023; Kolodiazhnyi et al., 2024) have recently emerged. Xiang et al. (2023) and Henrich et al. (2023) directly applied such networks to forest point clouds, demonstrating their substantial potential. However, tree species classification and individual tree segmentation are inherently complementary tasks, and the limited information provided by single-modality data struggles to meet the requirements of both tasks simultaneously.

*2.2.2. Multi-modal methods*

Recent studies have explored multi-modal approaches for tree species classification and individual tree segmentation, demonstrating the effectiveness of data complementarity in improving both tasks (Liu et al., 2017; Hartling et al., 2021; Qin et al., 2022). As for DL-based methods, Silvi-Net(Briechle et al., 2021) utilizes point clouds and multispectral images to train a dual-branch CNN network, achieving higher species classification accuracy than PointNet++ (Qi et al., 2017b) across two study regions. Liu et al. (2023a) proposed a multi-modal classification framework, TSCMDL, which directly extracts and fuses features from raw point clouds and images, thereby avoiding the information loss introduced by projecting point clouds into image. ShadowSense (Kapil et al., 2024) adopts a self-supervised training strategy to extract features without source domain annotations, combining complementary information from RGB and TIR images to enhance tree crown detection accuracy. Several methods (Zhang et al., 2023; Cao et al., 2025) have also been developed to process data collected by MMS in urban areas. However, these methods focus exclusively on either semantic segmentation or instance segmentation of non-tree traffic elements, such as vehicles, streetlights, and pedestrians.

In summary, DL-based algorithms have made significant advancements in the fields of tree species classification and individual tree segmentation. However, research on multi-modal DL approaches remains in its early stages, due to the limited scale and uni-modal nature of existing datasets, particularly in urban scenarios. To address this gap, we present WHU-STree, a multi-modal benchmark dataset for street tree inventory. With its large-scale, richly-annotated, and multi-modal characteristics, WHU-STree is expected to facilitate further advancements in the field.

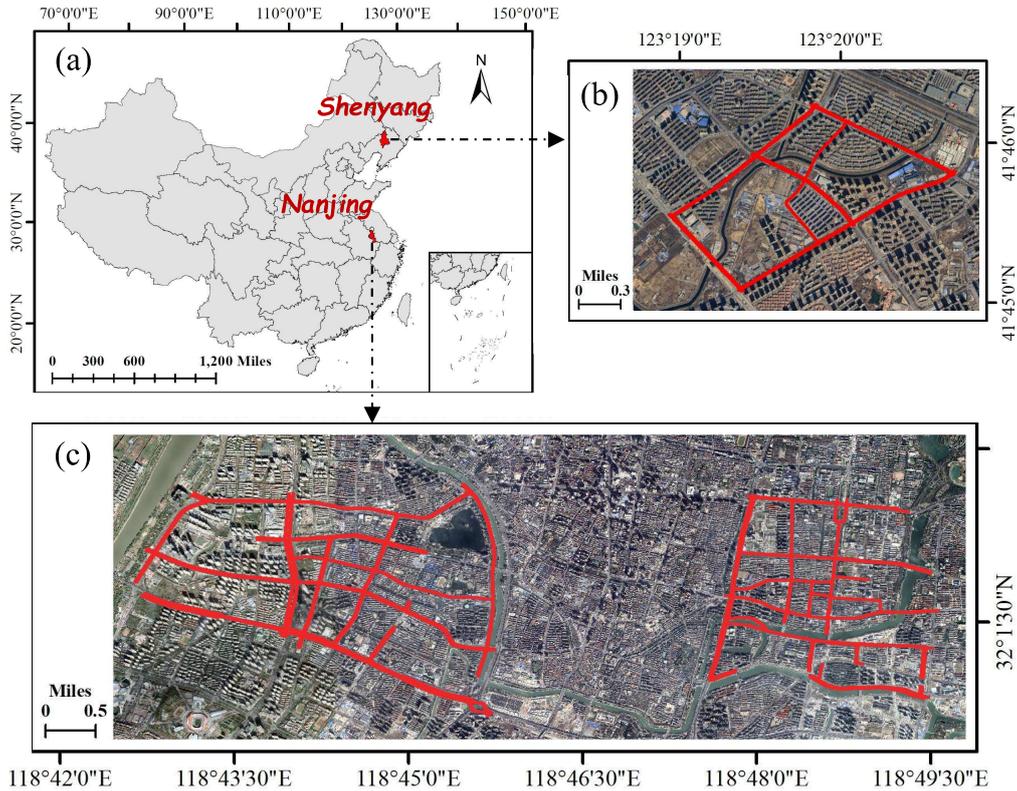

Fig. 2. Data acquisition region of WHU-STree dataset. (a) Location of Nanjing city and Shenyang city. (b) The road coverage of WHU-STree-SY. (c) The road coveragre of WHU-STree-NJ.

## 3. The WHU-STree Dataset

In this section, we introduce the construction process of WHU-STree and analyze the characteristics of the dataset.

*3.1. Data acquisition*

We collected data in Nanjing and Shenyang to construct the dataset (Fig. 2). Located approximately 970 km apart (straight-line distance) and falling within distinct climate zones, the two cities differ significantly in tree species, morphology, and street environments. The Nanjing dataset, designated as WHU-STree-NJ, was collected using the Hiscan-Z MMS in 2020 and 2021. The Shenyang dataset, referred to as WHU-STree-SY, was acquired in 2021 using the Alpha3D system.

Table 2
Parameters of the MMS equipment.

| Parameter | Hiscan-Z | Alpha3D |
|---|---|---|
| Maximum scanning distance | 119m | 420m |
| Laser emission frequency | 1010k pts/s | 1000k pts/s |
| Scanning frequency | 200Hz | 250Hz |
| Ranging accuracy | 9mm | 5mm |
| Panoramic camera resolution | 8192 × 4096 | 8192 × 4096 |

Both the Hiscan-Z and Alpha3D systems share similar configurations, integrating a laser scanner, a GNSS receiver, an inertial measurement unit (IMU), and a 360° panoramic camera, all synchronized by a central control module. Mounted on a rigid platform for convenient vehicle installation, these systems efficiently capture high-precision positioning and orientation system (POS) data, dense point clouds, and high-definition panoramic imagery during high-speed operation. The detailed specifications are listed in Table 2. Following data acquisition, the WHU-STree-NJ dataset comprises mobile mapping data covering approximately 59 km across 31 roads, while the WHU-STree-SY dataset covers about 7 km across 8 roads.

*3.2. Data processing and annotation*

**Preprocessing.** The raw mobile mapping dataset contains dense point clouds (averaging 616 and 1377 points/m$^2$ for Nanjing and Shenyang, respectively), panoramic images, and POS data. To improve annotation efficiency, preprocessing was applied. (1) Downsampling: voxel-based filtering (0.05 m voxel size) was used to reduce data volume. (2) Adaptive Partitioning: segments were divided at road intersections rather than fixed intervals to maintain continuity of road-side features (e.g., avoiding tree fragmentation). (3) Outlier Removal: Statistical filtering was applied to eliminate noise points, followed by ground point removal using an efficient ground filtering method (Zhang et al., 2016).

**Instances and species annotation.** Tree instances were annotated by trained personnel. For overlapping trees with ambiguous boundaries, anno-tators delineated tree crowns based on morphological characteristics, ensuring each instance maintains reasonable 3D structure representation. Subsequently, 50 species were identified in WHU-STree-NJ through integrated analysis of point clouds, panoramic imagery, and field surveys. For convenience, all tree species in the following text are abbreviated as listed in Appendix Table B.1. By using the extrinsic parameter at the moment of image exposure to register panoramic images with point clouds, 3D annotations were directly projected onto 2D image planes, eliminating the need for separate manual annotation of panoramic imagery.

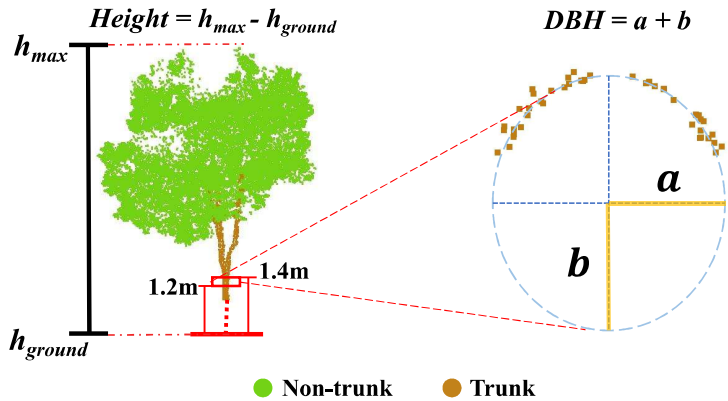

Fig. 3. Tree morphological parameter calculation based on point clouds.

**Morphological parameters calculation.** Tree positions were determined by the trunk's bottom center derived from annotated point clouds. As shown in Fig. 3, tree height was derived by subtracting the ground elevation ($h_{ground}$) from the maximum elevation of tree points ($h_{max}$). For diameter at breast height(DBH), following Weiser et al. (2022) , we estimated its value using tree point clouds to circumvent time-consuming and labor-intensive field measurements. Specifically, we first trained a classifier based on MinkNet(Choy et al., 2019) using annotations from the FOR-instance (Puliti et al., 2023) dataset to extract trunk segments from individual trees. Then, ellipse fitting was performed to horizontally sliced trunk point clouds at 1.2–1.4 m height to derive the major and minor axes, with their mean recorded as DBH. For tree instances with severe point loss, field-measured DBH values were incorporated instead.

Table 3
Quantitative descriptions of WHU-STree dataset.

| Datasets | Road length (km) | Data size (GB) | Trees | Species | Images |
|---|---|---|---|---|---|
| WHU-STree-NJ | 59.182 | 92.9 | 18343 | 50 | 10907 |
| WHU-STree-SY | 7.062 | 15.3 | 2664 | - | 1286 |
| Total | 66.244 | 108.2 | 21007 | 50 | 12193 |

*3.3. Data split and statistics*

After processing, the WHU-STree dataset comprises a total of 21,007 individual tree instances across two cities, with annotations for 50 tree species in the WHU-STree-NJ subset. Detailed quantitative information is presented in Table 3. Examples of the annotated data are shown in Fig. 4.

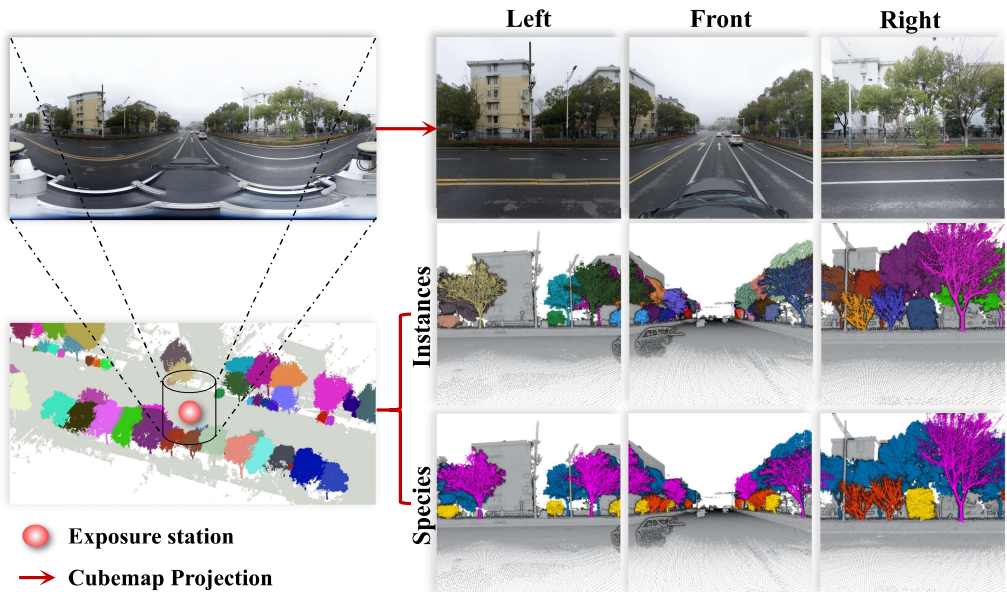

Fig. 4. Examples of our WHU-STree dataset. Instance annotations are assigned random colors for display. Tree species annotations are assigned distinct colors according to the following scheme: CC (blue), OF (yellow), ZS (purple), PS (red).

Nanjing and Shenyang are located in distinct climatic zones, leading to notable differences in street tree characteristics. We visualized the distribution of morphological parameters (i.e., height and DBH) in the WHU-STree dataset using violin plots, as shown in Fig. 5. WHU-STree-NJ exhibits broad height (predominantly 10-20m, occasionally exceeding 30m) and DBH distributions (mostly 0.2-0.5m), indicating robust growth likely attributable to favorable climatic conditions. Conversely, WHU-STree-SY shows more concentrated patterns, with heights primarily between 5-15m and DBH mostly below 0.3m. Overall, street trees in WHU-STree-NJ demonstrate a clear advantage in both height and DBH, reflecting distinct differences in urban tree management and environmental conditions between sub-dataset coverage zones.

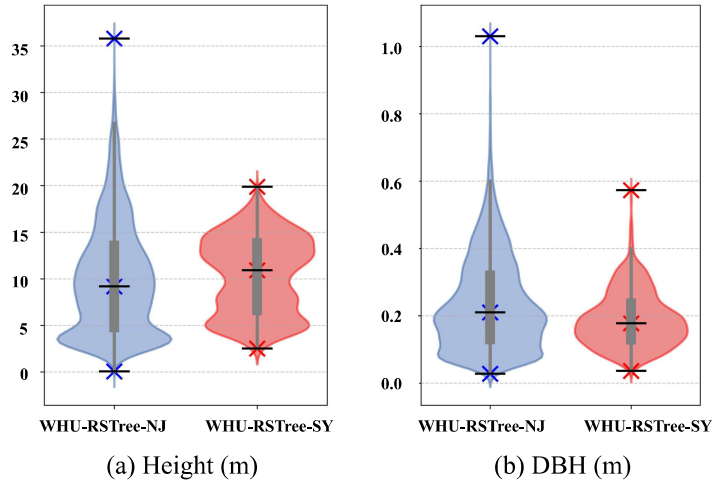

| (a) Height (m) | (b) DBH (m) |

Fig. 5. Violin plots for DBH and height of trees in WHU-STree dataset.

Urban trees exhibit significant spatial heterogeneity within sub-datasets. Planting density, individual size, and species composition differ markedly between streets. As exemplified by the WHU-STree-NJ dataset, densities range from very sparse (less than one tree per 100m) to high (nearly one tree per meter). Planned planting typically yields homogeneous species assemblages per street, resulting in rare species limited to few locations while prevalent species are widely distributed citywide. Beyond spatial distribution, notable disparities exist in per-species abundance and intraspecific morphological variation, as illustrated in Fig. 6.

The extensive imbalance both across and within datasets increases the complexity of analyzing street trees and places high demands on the robustness and adaptability of algorithms. Nevertheless, it significantly enriches the diversity of our dataset, thereby enhancing its scientific value.

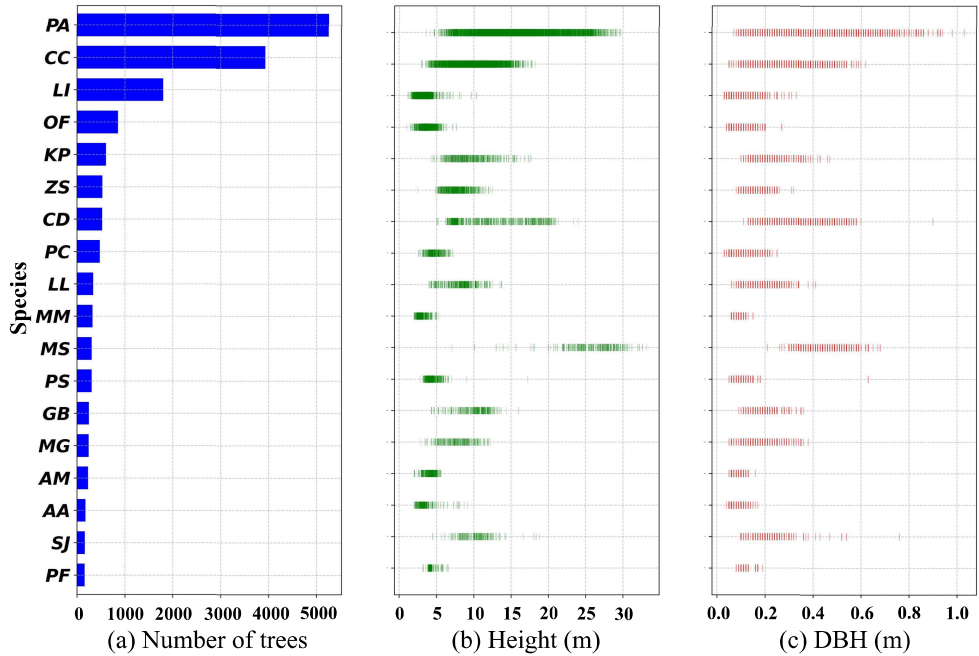

Fig. 6. Tree species frequency, height distributions and DBH distributions in the WHU-STree-NJ.

Based on the dataset characteristics analyzed above, we implemented two distinct data splitting strategies. (1) Class-balanced split: considering the significant impact of class distribution on classification tasks, we uniformly sample specific road sections to form the test set. This split maintains a roughly consistent class ratio between the training and test sets, ensuring a relatively balanced test setting. (2) Cross-city split: in different cities, the substantial variations in tree shape and size require tree segmentation algorithms to exhibit robust performance and strong generalization capabilities. Thus, the cross-city split is proposed for systematic evaluation of algorithmic generalizability.

The schematic diagram depicting the two splitting methods is presented in Fig. 7. The first splitting method is suitable for both tree species classification and individual tree segmentation algorithms. The second specifically focuses on evaluating the cross-domain generalization capability of individual tree segmentation algorithms in diverse urban environments.

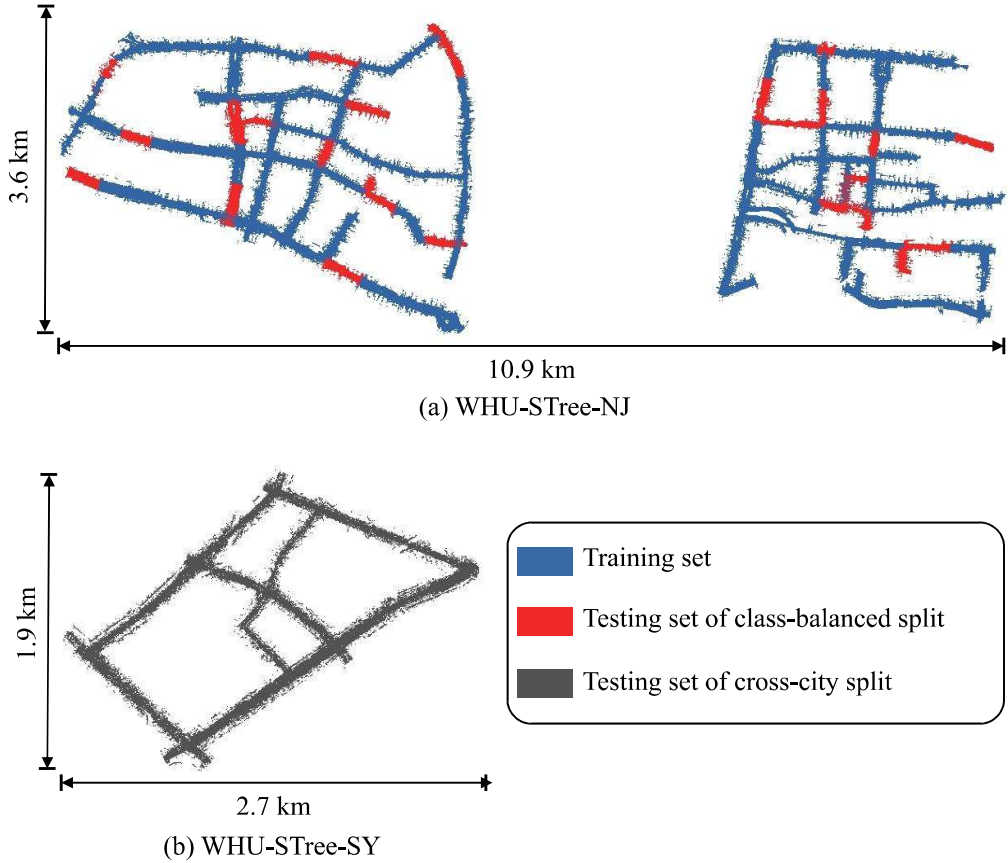

Fig. 7. Data split of the WHU-STree Dataset.

*3.4. Unique advantages*

Compared to existing datasets, the WHU-STree dataset offers several key advantages: cross-city coverage, rich annotations, and multi-modal data, enabling support for diverse task settings. These strengths make the dataset highly valuable for urban street tree analysis, environmental monitoring, and the training and evaluation of DL-based models.

**Cross-city coverage.** The WHU-STree dataset encompasses two geographically distant cities, Nanjing and Shenyang, which represent distinct geographical and climatic regions in China with varying urban planning, tree distribution, and environmental characteristics. Consequently, this dataset enables the evaluation of model performance across diverse urban environ-ments, providing valuable insights into their adaptability and generalization capabilities across regions (Chen et al., 2022). Furthermore, the dataset includes 66 km of point cloud data and over 12,000 panoramic images, with annotations for 21,007 tree instances and 50 tree species. In terms of scale, it significantly surpasses existing tree segmentation datasets, and its number of tree instances is comparable to that of the largest currently available tree species classification dataset (Puliti et al., 2024). This large-scale and diverse dataset offers abundant training samples, helping to improve DL-based models' generalization and robustness. In summary, WHU-STree not only serves as a benchmark for evaluating generalization but also provides substantial data support for enhancing algorithmic generalization capabilities.

**Rich annotations.** The proposed dataset provides extensive and detailed annotations, encompassing not only precise delineations of individual tree instances with accurate localizations but also species information and comprehensive morphological parameters. These rich annotations establish the dataset as an excellent benchmark for various research tasks, including tree species classification, individual tree segmentation, and morphological parameter estimation. Furthermore, the diverse annotation types support advanced research, such as the development of multi-task learning models that simultaneously perform classification, segmentation, and parameter estimation, thereby facilitating cross-task knowledge transfer and promoting a deeper understanding of the relationships among semantic, structural, and morphological features (Zhou et al., 2025).

**Multi-modal data.** This dataset stands out for the multi-modal integration of point cloud and panoramic imagery. The point cloud modality provides precise 3D structural data, such as tree height, crown shape, and spatial distribution, which are vital for morphological analysis and instance localization. In parallel, panoramic images offer high-resolution visual details, including texture, color, and environmental context, which are critical for species identification and semantic understanding. By fusing these complementary modalities, the dataset facilitates the development of algorithms that jointly exploit 3D geometry and 2D appearance, thereby enhancing model comprehensiveness and robustness. We hope that the WHU-STree will promote progress in multi-modal learning, fostering the development of sophisticated methods for urban street tree asset management.

**Multi-task support.** With its characteristics of cross-city coverage, rich annotations, and multi-modal data, the WHU-STree dataset offers substantial flexibility in task design. It supports a variety of configurations, including single-modality or multi-modality input, single-task or multi-task learning, and cross-region or within-region evaluations. This flexibility accommodates diverse research tasks related to urban street tree inventories and fosters methodological innovation. Specifically, the dataset enables fundamental tasks such as tree species classification, individual tree segmentation, and morphological parameter estimation based on either point cloud or image data. Beyond these, WHU-STree facilitates the exploration of advanced learning paradigms, including multi-modal data fusion, domain adaptation/generalization, and the multi-task learning. Moreover, the richness and diversity of the dataset make it possible to define new tasks and benchmarks, such as estimating 3D tree morphological parameters directly from 2D images—a challenging yet highly promising direction. Success in such a task would enable the use of crowd-sourced street view imagery to establish a street tree inventory at significantly reduced costs. We believe that the release of WHU-STree will further advance research in MMS-based urban street tree management and contribute to the construction of smart ecological cities.

## 4. Benchmark Experiments

For tree species classification and individual tree segmentation, we carefully selected several representative methods, including both uni-modal and multi-modal approaches, to benchmark the proposed WHU-STree dataset. Based on the results, we provide in-depth analysis and valuable insights. It should be noted that during the experiment, we group tree species with less than 150 instances under the label *"Others"*.

### 4.1. Tree species classification baselines

For tree species classification using point clouds, established approaches primarily adopt three paradigms: voxel-based, point-based, and transformer-based methods. We selected one representative baseline from each category. Additionally, we reproduced TSCMDL based on the model description in the original paper and employed it as the multi-modal baseline. The selected baselines are:

- MinkNet (Choy et al., 2019) employs 3D sparse convolutions on voxelized point clouds using the MinkowskiEngine. By representing input

data as sparse tensors, the inefficiency and high memory consumption associated with dense representations in 3D laser scanning can be avoided.

- PointMLP (Ma et al., 2022) introduces a pure residual MLP network, which integrates no sophisticated local geometrical extractors but still performs very competitively. It developed a lightweight geometric affine module to tackle the decrease of accuracy and stability caused by sparse and irregular geometric structures in local regions.

- PTv2 (Wu et al., 2022) improves PTv1 through three core innovations: Grouped Vector Attention (GVA) reduces parameters by grouping channels with shared weights; Position Encoding Multiplier (PEM) adds a multiplicative term to enhance 3D spatial awareness; Partition-based pooling uses grid partitions for efficient feature aggregation.

- TSCMDL (Liu et al., 2023a) aims to develop a multi-modal framework that integrates 2D and 3D features for tree species classification. This framework employs ResNet (He et al., 2016) and PointMLP (Ma et al., 2022) to extract image features and point cloud features, respectively, and then uses the fused features generated by concatenation for final classification.

*4.2. Tree segmentation baselines*

Leveraging the comprehensive annotations available in WHU-STree, we extend the tree instance segmentation task to simultaneously segment individual trees and classify their species. This is achieved by modifying the classification head of standard instance segmentation models. The recent predominance of clustering-based and transformer-based methodologies in instance segmentation research prompts our adoption of SoftGroup (Vu et al., 2022) and SPFormer (Sun et al., 2023) as representative baselines. We further incorporate SegmentAnyTree (Wielgosz et al., 2024), an extension of PointGroup (Jiang et al., 2020) specifically optimized for individual tree segmentation. Additionally, we evaluate LCPS (Zhang et al., 2023), a multi-modal panoptic segmentation algorithm that integrates point clouds and RGB images, to investigate the potential benefits of leveraging multi-modal data for tree segmentation.

- SegmentAnyTree (Wielgosz et al., 2024) uses a bottom-up panonptic segmentation pipline consisting of a shared feature extractor followed by three parallel prediction branches. The first branch is dedicated to semantic segmentation. The remaining two branches implement two complementary instance clustering strategies—centroid offset regression and embedding-based contrastive learning—to address the complexity of tree segmentation in densely packed forests.

- SoftGroup (Vu et al., 2022) aims at addressing semantic prediction errors by retaining soft semantic probabilities during clustering. Instead of relying on hard semantic labels, it allows cross-category clustering and refines instance proposals using a top-down network to correct misclassified points. This soft-label strategy improves robustness to noisy semantics.

- SPFormer (Sun et al., 2023) proposes an end-to-end instance segmentation method that unifies superpoint representation and transformer-based query decoding. It aggregates point features into superpoints via average pooling and employs learnable queries to decode instance masks through cross-attention, eliminating reliance on error-prone intermediate grouping steps.

- LCPS (Zhang et al., 2023) introduces the first LiDAR-camera fusion framework for 3D panoptic segmentation, featuring a three-stage alignment strategy: 1) Asynchronous Compensation Pixel Alignment (ACPA) addresses coordinate misalignment from sensor asynchronicity via ego-motion compensation; 2) Semantic-Aware Region Alignment (SARA) extends point-to-pixel mapping to semantic regions using class activation maps; 3) Point-to-Voxel Propagation (PVP) employs local attention to propagate fused features across the entire point cloud.

*4.3. Evaluation metrics*

As for tree species classification, the evaluation metrics are Overall Accuracy (OA), and Intersection over Union (IoU):

$$OA = \frac{\sum_{k=1}^{K} TP_k}{GT}, \qquad (1)$$

$$IoU_k = \frac{TP_k}{TP_k + FP_k + FN_k}, \qquad (2)$$

where $TP_k$, $FP_k$, $FN_k$ correspond to the true positives, false positives, and false negatives values of the $k$th tree species, and $GT$ denotes the total number of ground truth instances.

For tree segmentation baselines, we follow the evaluation protocol of Wang et al. (2019) and adopt mean class precision (mPrec), mean class recall (mRec), mean class coverage (mCov), and mean class-weighted coverage (mWCov) as species-specific metrics. Among these, mPrec and mRec are calculated using a IoU threshold of 0.5. However, these metrics inherently reflect the combined performance of both segmentation and classification. To enable targeted evaluation of species-agnostic tree segmentation, we introduce a set of supplementary metrics, including Detection rate (Det), Omission rate (Omi), Commission rate (Com), and F1 :

$$Detection\ rate = \frac{\sum_{k=1}^{K} TP_k}{GT}, \qquad (3)$$

$$Omission\ rate = \frac{\sum_{k=1}^{K} FN_k}{GT}, \qquad (4)$$

$$Commission\ rate = \frac{\sum_{k=1}^{K} FP_k}{PT}, \qquad (5)$$

$$F1 = \frac{2 \cdot Precision \cdot Recall}{Precision + Recall}, \qquad (6)$$

where $PT$ refers to the predicted trees, and the threshold for determining $TP$ is set to an IoU of 0.5.

*4.4. Results and discussion*

*4.4.1. Experiments for tree species classification*

Table 4 presents three point cloud-based tree species classification methods (MinkNet (Choy et al., 2019), PointMLP (Ma et al., 2022), PTv2 (Wu et al., 2022)) and one multi-modal method (TSCMDL (Liu et al., 2023a)) that integrates both image and point cloud information. The voxel-based MinkNet exhibits the lowest accuracy, primarily due to information loss during voxel quantization, which discards fine-grained structural details. In contrast, the point-based PointMLP achieves higher performance, while PTv2 leverages self-attention mechanisms to attain state-of-the-art results. TSCMDL employs PointMLP and ResNet (He et al., 2016) to extract features from point clouds and images, respectively. The fusion of these two modalities yields significant performance improvements over PointMLP alone, demonstrating the effectiveness of multi-modal fusion. The visualization of the confusion matrix highlights the categories that are easily confused or misclassified, as shown in Appendix Fig. D.3 - Fig. D.6.

Table 4
Quantitative results of tree species classification benchmark on WHU-STree-NJ.

| Method | mIoU↑(%) | OA↑(%) |
|---|---|---|
| MinkNet(Choy et al., 2019) | 49.94 | 81.67 |
| PointMLP(Ma et al., 2022) | 58.00 | 84.51 |
| PTv2(Wu et al., 2022) | **64.09** | **87.99** |
| TSCMDL(Liu et al., 2023a) | 60.49 | 86.20 |

Further analysis of the samples reveals that some categories are difficult to distinguish from the point cloud, while the image can serve as a valuable supplementary source of information for classification, as demonstrated by the categories LI and AM in Fig. 8a and Fig. 8b. In addition, there is very little inter-class variance between some tree species, such as LL and CC in Fig. 8c and Fig. 8d. Even with multi-modal data, they are difficult to distinguish, which presents a significant challenge for algorithm design.

**Key findings.** PTv2 achieves optimal performance, validating the effectiveness of long-range feature modeling for species classification. Moreover, compared to single-modality approaches, the multi-modal fusion strategy yields superior results with 2.49% improvements in mIoU, highlighting the importance of leveraging complementary data. However, despite incorporating supplementary image information, distinguishing specific tree species remains a significant challenge, as evidenced by the highest obtained mIoU of just 64.09%.

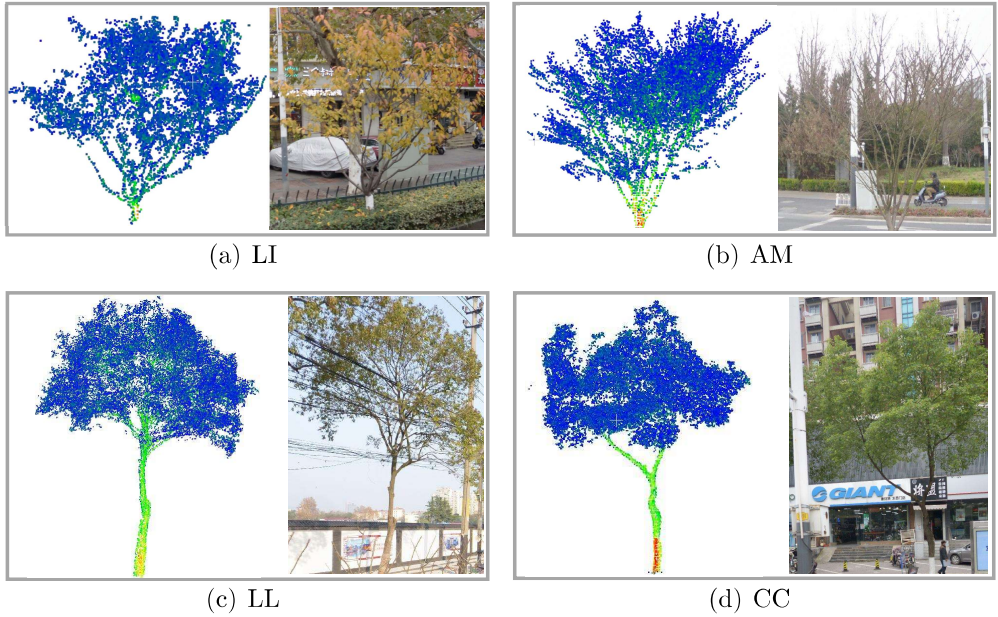

Fig. 8. Visualization of representative tree species in point clouds and images. The left of each subfigure displays the tree point clouds color-coded by intensity, while the right side presents its corresponding image.

*4.4.2. Experiments for tree segmentation*

Table 5 presents the quantitative evaluation of three uni-modal and one multi-modal algorithms for species-agnostic tree segmentation on the WHU-STree-NJ and WHU-STree-SY datasets. On WHU-STree-NJ, the transformer-based SPFormer achieved the highest Det (93.2%), benefiting from its end-to-end attention mechanism that models complex spatial relationships without intermediate error accumulation. However, it exhibited a notably higher Com (26.6%), mainly due to shrub misclassification. Among clustering-based methods, SoftGroup performed the weakest, struggling with dense or overlapping trees due to offset prediction. SegmentAnyTree, incorporating contrastive-loss-trained embedding features for clustering, achieved the highest F1-score (86.8%) by effectively segmenting challenging trees. LCPS projects features to Bird's-Eye-View (BEV), incurring vertical information loss, which limits the differentiation of suppressed subcanopy trees, yielding a suboptimal Det (83.0%). In cross-city evaluations on the WHU-STree-SY dataset, all algorithms demonstrated robust performance. This consistency can be attributed to the comprehensive nature of our dataset, which spans diverse scenes, tree species, and morphologies, providing a solid foundation for real-world algorithmic applications.

Table 6 presents results for tree segmentation integrated with species classification on the WHU-STree-NJ dataset. Although satisfactory performance was achieved for specific categories (e.g., PA, CC), overall effectiveness remains suboptimal. Species classification within instance segmentation is inherently more challenging than classification with isolated trees. In clustering-based instance segmentation algorithms, classification results are typically determined by the majority vote of per-point predictions within an instance mask. However, per-point predictions lack a holistic understanding of structure, and segmentation errors compromise classification accuracy. SPFormer achieves state-of-the-art performance (57.8% in F1-score) due to its superpoint-based mask classification paradigm. These geometrically derived superpoints adaptively preserve homogeneous clusters, overcoming the limitations of uniform voxels in feature discriminability. Despite utilizing RGB images, LCPS underperforms compared to the uni-modal SPFormer. Beyond inferior segmentation, its cylindrical voxel representation fails to leverage the advantages of sparse, unevenly distributed point cloud data.

Qualitative results on the WHU-STree-NJ and WHU-STree-SY are shown in Appendix Fig. D.7 and Appendix Fig. D.8. SegmentAnyTree demonstrates fewer errors and superior performance in resolving confusion with shrubs. SoftGroup exhibits widespread errors and struggles with segmenting connected canopies. SPFormer misidentifies certain shrubs as trees, leading to a higher Com. LCPS performs the worst when segmenting understory trees beneath tall canopies. Furthermore, we highlight three representative challenging scenarios (Fig. 9) to emphasize the research value of our dataset and inspire future algorithmic studies. Fig. 9a shows mis-segmentation caused by interference from urban road furniture, especially they are adjacent to trees. While such errors are often minimal in point count and segmentation metrics, they can critically impact downstream tasks (e.g., parameter estimation). Fig. 9b depicts severe crown overlap, where dense canopies lack discernible boundaries, leading to under-segmentation. Fig. 9c illustrates errors from block merging failures, characterized by arcuate boundaries due to cylindrical block sampling. While the patch-then-merge strategy is essential for large-scale processing under computational constraints, it introduces challenging prediction fusion issues. Current heuristic-based merging techniques lack the flexibility to handle all cases.

Table 5
Quantitative results of species-agnostic tree segmentation benchmark on WHU-STree.

| Method | WHU-STree-NJ | | | | WHU-STree-SY | | | |
| --- | --- | --- | --- | --- | --- | --- | --- | --- |
| | Det↑(%) | Omi↓(%) | Com↓(%) | F1↑(%) | Det↑(%) | Omi↓(%) | Com↓(%) | F1↑(%) |
| SegmentAnyTree(Wielgosz et al., 2024) | 87.6 | 12.4 | **14.1** | 86.8 | 86.0 | 14.0 | 12.6 | 86.7 |
| SoftGroup(Vu et al., 2022) | 59.0 | 41.0 | 33.0 | 62.7 | 79.9 | 20.1 | 20.1 | 79.9 |
| SPFormer(Sun et al., 2023) | **93.2** | **6.8** | 26.6 | 82.2 | **87.4** | **12.6** | 17.5 | 84.9 |
| LCPS (Zhang et al., 2023) | 75.6 | 24.4 | 17.8 | 78.8 | 83.0 | 17.0 | **7.8** | **87.4** |

Table 6
Quantitative results of species-specific tree segmentation benchmark on WHU-STree-NJ.

| Method | mPrec↑(%) | mRec↑(%) | mCov↑(%) | mWCov↑(%) | F1↑(%) |
| --- | --- | --- | --- | --- | --- |
| SegmentAnyTree(Wielgosz et al., 2024) | **56.9** | 46.5 | 43.2 | 46.8 | 51.2 |
| SoftGroup(Vu et al., 2022) | 36.0 | 31.1 | 31.4 | 31.6 | 33.4 |
| SPFomer(Sun et al., 2023) | 52.8 | **63.9** | **59.4** | **61.8** | **57.8** |
| LCPS(Zhang et al., 2023) | 41.9 | 38.0 | 35.3 | 35.0 | 39.9 |

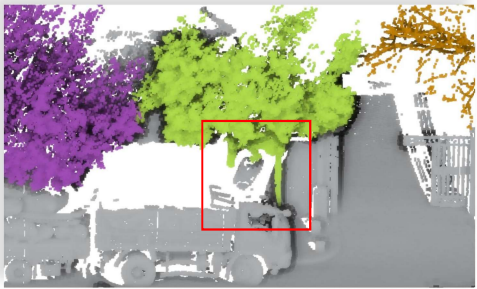 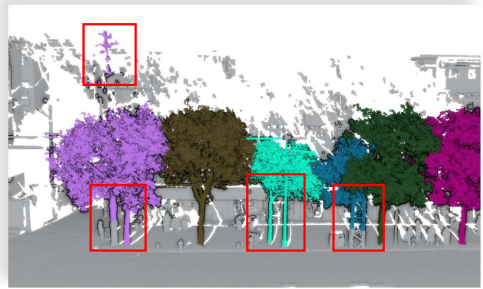

(a) Furniture interference

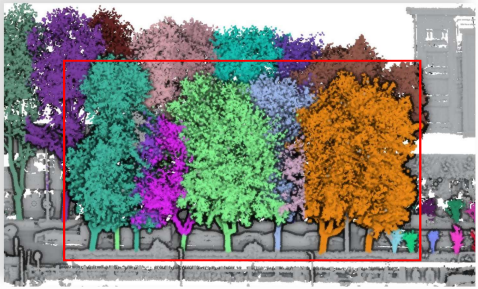 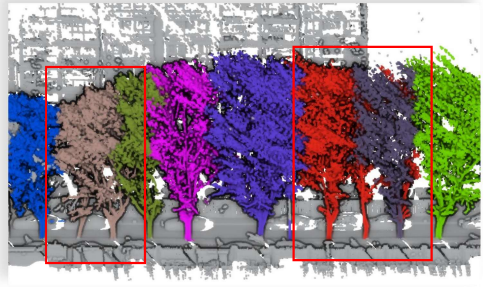

(b) Crown overlap

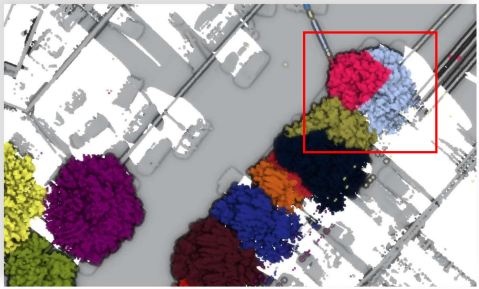 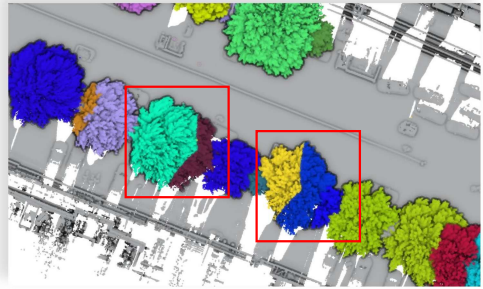

(c) Block merging failure

Fig. 9. Representative challenging scenarios for tree segmentation.

**Key findings.** Current algorithms demonstrate stable performance on the segmentation task; however, their integrated segmentation-classification performance is notably poor, with the highest F1-score reaching only 57.8%. Additionally, transformer-based paradigms clearly outperform clustering-based paradigms across key evaluation metrics. The introduction of the image modality provides improvements for both species-agnostic and species-specific metrics, with the spectral and texture information from images proving particularly helpful for distinguishing between species.

## 5. Challenges and Outlook

### 5.1. Fusion of point clouds and images

Uni-modal data limitations drive multi-modal fusion research, yet multi-modal integration remains scarce in individual tree segmentation due to lacking datasets. WHU-STree addresses this by providing densely aligned point clouds and panoramic images with calibrated projection parameters. This enables precise geometric alignment between 3D points and 2D pixels, establishing a solid foundation for multi-modal urban street tree analysis.

Table 7
Experiment results for validating the performance improvement achieved through the incorporation of image modality.

| Method | WHU-STree-NJ | | | | WHU-STree-SY | | | |
|---|---|---|---|---|---|---|---|---|
| (Species-agnostic) | Det↑(%) | Omi↓(%) | Com↓(%) | F1↑(%) | Det↑(%) | Omi↓(%) | Com↓(%) | F1↑(%) |
| LCPS(3D Only) (Zhang et al., 2023) | 74.0 | 26.0 | 25.3 | 74.3 | 76.3 | 23.7 | 8.6 | 83.2 |
| LCPS (Zhang et al., 2023) | 75.6 | 24.4 | 17.8 | 78.8 | 83.0 | 17.0 | 7.8 | 87.4 |
| (Species-specific) | mPrec↑(%) | mRec↑(%) | mCov↑(%) | mWCov↑(%) | F1↑(%) | mPrec↑(%) | mRec↑(%) | mCov↑(%) | mWCov↑(%) | F1↑(%) |
| LCPS(3D Only) (Zhang et al., 2023) | 26.0 | 35.5 | 24.5 | 26.5 | 30.0 | - | - | - | - | - |
| LCPS (Zhang et al., 2023) | 41.9 | 38.0 | 35.3 | 35.0 | 39.9 | - | - | - | - | - |

We adapt the LCPS as a baseline for multi-modal tree segmentation. To rigorously asses the contribution of image modality while maintaining architectural consistency, we conducted ablation studies using 3D-only LCPS configurations. Experimental results are presented in Table 7. The LCPS demonstrates notable improvements over the 3D-only baseline in species-agnostic individual tree segmentation, achieving increases of +1.6% in mPrec and +4.5% in F1-score. The most substantial enhancement is observed in Com reduction, with a 7.5% decrease. These performance improvements demonstrate that integrating the image modality can significantly enhance the robustness in complex urban street tree scenarios. When it comes to species-specific results, LCPS consistently outperforms its 3D-only baseline across all metrics: mPrec increases by +6.9%, mRec by +12.0%, mCov by +10.8%, mWCov by +8.5%, and F1-score by +9.9%. These results underscore the critical role of image-derived texture and color features in enhancing species classification accuracy.

Despite improvements, performance remains impractical for deployment. Nevertheless, it represents a valuable exploration that warrants further investigation. Several critical challenges persist, including projection errors and multi-view inconsistencies, as well as spatiotemporal misalignments(Yi et al., 2025). These inherent limitations require further in-depth study (Han et al., 2024). To this end, we advocate the development of task-specific multi-modal fusion strategies tailored to street tree inventory. We anticipate that the WHU-STree dataset will promote the advancement of multi-modal approaches in urban street tree management, fostering both broader adoption and deeper exploration.

### 5.2. Collaboration of multi-task

Multi-task learning integrates concurrent tasks into end-to-end frameworks, streamlining complex processing (e.g., UniAD (Hu et al., 2023), ReasonNet (Shao et al., 2023)) and enabling shared representation learning for mutual performance enhancement (e.g., WHU-synthetic (Zhou et al., 2025), PatchAugNet (Zou et al., 2023)). For street tree inventory, tree species classification and individual tree segmentation are critical yet typically addressed separately. This disjointed pipeline increases complexity, propagates errors, and complicates evaluation. Crucially, existing datasets hinder exploration of multi-task collaboration. The proposed WHU-STree addresses this gap by simultaneously providing tree instances and species annotations. This enables the development of end-to-end multi-task learning frameworks capable of directly predicting tree instances with associated species labels, paving the way for more efficient and robust urban tree inventory systems.

Table 8
Experiment results for validating the effectiveness of collaboration of tree species classification and individual tree segmentation.

| Method | WHU-STree-NJ | | | | WHU-STree-SY | | | |
|---|---|---|---|---|---|---|---|---|
| | Det↑(%) | Omi↓(%) | Com↓(%) | F1↑(%) | Det↑(%) | Omi↓(%) | Com↓(%) | F1↑(%) |
| SPFormer (Sun et al., 2023) (w/o species) | 93.7 | 6.3 | 45.2 | 69.2 | 92.5 | 7.5 | 33.4 | 77.5 |
| SPFormer (Sun et al., 2023) (w/ species) | 93.2 | 6.8 | 26.6 | 82.2 | 87.4 | 12.6 | 17.5 | 84.9 |

To further validate the effectiveness of multi-task collaboration, we conducted preliminary experiments. As previously described, in the tree segmentation baseline, we tasked the semantic or category branch with distinguishing tree species. For comparison, we modified the baseline to differentiate only between trees and non-trees and assessed the species-agnostic performance of both approaches. The results are presented in Table 8. On the WHU-STree-NJ dataset, distinguishing tree species compared to not distinguishing them results in a comparable Det (93.7% vs. 93.2%) but a higher F1-score (82.2% vs. 69.2%). This indicates that incorporating species classification enhances the network's ability to capture fine-grained variations among trees, indirectly improving the overall understanding of the "tree" concept and yielding more precise results. In cross-domain tests, while F1-score shows some improvement by +7.5 %, Det slightly declines by -5.1 %. We hypothesize that while species classification facilitates learning of species-specific shape representations, inter-species variations across cities may limit detection generalizability.

These preliminary findings demonstrate the effectiveness and substantial potential of multi-task collaboration in street tree inventory. Moving forward, we plan to incorporate additional annotations, including canopy width, support structure ratio, and health status, to support more comprehensive multi-task collaborative frameworks. We encourage researchers to utilize WHU-STree to further investigate and design advanced multi-task frameworks, thereby unlocking enhanced capabilities for street tree inventory.

5.3. Cross-domain generalization

DL-based methods often exhibit bias toward the distribution of the training data. In real-world applications, variations in object appearance or ac-quisition equipment typically lead to domain shifts, resulting in poor out-of-distribution (OOD) performance (Saltori et al., 2022; Chen et al., 2024b). Therefore, generalization is a critical factor for ensuring the practical applicability of algorithms. As previously discussed, WHU-STree-NJ and WHU-STree-SY differ substantially in point density, species composition, and tree morphology. To investigate generalization, we conducted cross-city experiments by training models on WHU-STree-NJ and evaluating them on WHU-STree-SY, focusing on the task of individual tree segmentation, with results presented in Table 5. Across four different baselines, performance did not show a noticeable decline and, in some cases, even slightly improved. This is likely attributable to the large volume and diversity of tree species and morphologies in WHU-STree-NJ, enabling the trained models to inherently possess a degree of generalization. Additionally, the use of voxel downsampling to a fixed resolution helped mitigate differences in point cloud density. Furthermore, the tree layout in WHU-STree-SY is more spacious, with less crown overlap and fewer shrub interferences, which reduces the complexity of individual tree segmentation. Notably, compared to the within-domain results on WHU-STree-NJ, the multi-modal method LCPS outperformed both SegmentAnyTree and SPFormer, achieving the highest F1 score of 87.4%. These results highlight the advantage of incorporating image modalities in alleviating domain shifts in point cloud data, and demonstrate the effectiveness of multi-modal fusion in improving robustness under heterogeneous urban conditions. We also expect this dataset to further advance research on cross-modal data domain adaptation and generalization (Jaritz et al., 2020).

To further support this line of work, we plan to incorporate species annotations for WHU-STree-SY in future releases, enabling robustness evaluation for tree species classification across domains. However, due to the combined effects of ecological and anthropogenic factors, tree species compositions vary significantly between cities. In this context, advancing research on open-vocabulary recognition (Radford et al., 2021; Liu et al., 2024) and novel class discovery (Fini et al., 2021; Riz et al., 2023) is particularly important. Moreover, we will continue to expand WHU-STree to cover additional cities, thereby supporting broader investigations into domain generalization in urban tree inventory tasks.

*5.4. Spatial pattern learning*

Unlike trees in natural forests, the spatial distribution and species composition of street trees exhibit strong anthropogenic characteristics. Street trees typically follow regular distribution patterns that adhere to planting standards established by local authorities, often arranged linearly along both sides of roads at fixed intervals (Messier et al., 2025). Species selection also reflects distinct function-oriented preferences, influenced by factors such as road hierarchy and functional zoning. For instance, arterial roads are often lined with tall canopy trees, while pedestrian streets tend to favor ornamental species. Understanding the spatial patterns of street trees holds dual significance for both automated inventory tasks and urban greening management. Incorporating street tree distribution patterns as prior knowledge can enhance segmentation and classification performance. In individual tree segmentation tasks, encoding spatial distribution patterns (e.g., linear arrangement along road networks, planting intervals, and topological relationships with surrounding infrastructure) can improve DL-based models' ability to distinguish trees from background points while mitigating over-segmentation and under-segmentation in complex scenarios. Similarly, species classification can benefit from spatial pattern learning. The function-oriented nature of street trees dictates distinct species preferences across different urban zones. Furthermore, trees in the same road typically maintain species uniformity or similarity. Therefore, spatial context-aware classifiers can leverage these characteristics as auxiliary cues to achieve more robust classification performance (Chi et al., 2025). Beyond fundamental analysis tasks, understanding spatial patterns proves crucial for effective monitoring and long-term management of street trees. It provides foundational data for assessing greening equity and ecological service efficiency, thereby informing rational urban greening planning (Martin et al., 2025).

In conclusion, we posit that spatial pattern learning represents a critical research frontier for future investigations. This approach not only enhances technical solutions for urban tree analysis but also establishes an important knowledge base for sustainable urban street trees management.

*5.5. MLLM for Street Tree Asset Management*

Urban street tree inventory encompasses a wide range of attributes, including location, species, morphological parameters, health condition, maintenance requirements (e.g., pruning), and risk assessment (e.g., potential branch fall hazards). Developing a comprehensive method capable of extracting and evaluating these attributes is pivotal for automating street tree asset management. However, attributes such as tree health, maintenance requirements, and risk assessment often demand expert knowledge or alignment with governmental policies for accurate interpretation. MLLM (Liu et al., 2023c; Yang et al., 2024a), with its ability to reason and interact across diverse data modalities, have shown remarkable progress in various vertical domains (Li et al., 2023a; Muhtar et al., 2024). Leveraging the multi-modal, richly annotated, and large-scale characteristics of the proposed WHU-STree dataset, we envision significantly enhancing the capability of MLLM to process and analyze urban street tree inventories. By integrating expert knowledge and governmental policies, MLLM can enable a seamless "perception-analysis-decision" pipeline for urban asset management. The ultimate goal is to establish a closed-loop urban street tree asset management framework powered by MLLM. In the perception phase, the model processes multi-modal data (point clouds, images, and potentially textual reports) to extract comprehensive tree attributes. In the analysis phase, it incorporates expert knowledge and policy guidelines to assess tree health, maintenance needs, and associated risks. Finally, in the decision phase, the model delivers actionable insights, such as maintenance schedules, risk mitigation plans, or policy-compliant management strategies, which can be queried through natural language interfaces. For instance, urban planners could ask, "Which trees exceed the permitted height limits and pose risks to overhead infrastructure?" and receive detailed, policy-aligned recommendations. This vision aligns with emerging trends in smart city applications, where multi-modal AI systems are increasingly used for infrastructure management.

In summary, we envision that by leveraging the WHU-STree dataset, integrating expert knowledge, and embedding governmental policies, MLLM can revolutionize urban street tree asset management. This approach not only enhances the accuracy and robustness of automated assessments but also ensures that management decisions are well-informed, compliant, and actionable, paving the way for sustainable and intelligent street tree asset management.

## 6. Conclusion

This study presents WHU-STree, a pioneering multi-modal dataset designed to address critical gaps in urban street tree inventory research. By integrating synchronized point clouds and high-resolution images collected across two geographically distinct cities, WHU-STree offers a comprehensive resource for advancing automated urban street tree inventory. The dataset encompasses 21,007 annotated tree instances with labels for 50 tree species and morphological parameters, enabling simultaneous support for tasks such as tree species classification, individual tree segmentation, and 3D morphological analysis. Its cross-city coverage, combined with rich annotations and multi-modal integration, sets a new benchmark for evaluating algorithmic generalization and robustness in real-world urban environments.

Benchmark experiments on WHU-STree demonstrates the superiority of multi-modal approaches over unimodal methods, particularly in reducing commission rate and enhancing classification accuracy through complementary 2D-3D feature fusion. The cross-city evaluation further highlights the adaptability of trained models to heterogeneous environments, underscoring the importance of domain-invariant representations for practical deployment. In addition, WHU-STree facilitates novel research directions, including spatial pattern analysis of human-influenced tree distributions and the integration of MLLM for holistic tree asset management.

Looking ahead, WHU-STree is poised to catalyze innovations in urban street tree asset management by bridging the gap between algorithmic research and real-world application. Future extensions could incorporate additional annotations (e.g., health status) and expand geographic coverage to further enhance its utility. This dataset contributes to the development of smarter, sustainable cities where the ecological, social, and economic benefits of urban trees are fully optimized.

## Acknowledgments

Our research has received funding from the National Key Research and Development Program of China, under Grant 2023YFF0725200.

## Appendix A. Supplementary figures

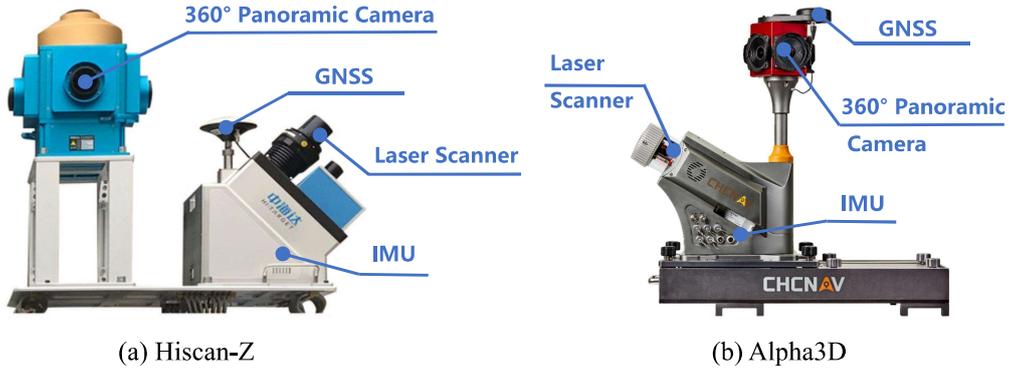

(a) Hiscan-Z  (b) Alpha3D

Fig. A.1. The MMS equipment which is used to collect data for building the WHU-STree.

## Appendix B. Abbreviations

Please see Table B.1.

## Appendix C. Implementation details

The input data processing pipeline for the tree species classification baselines is shown in Fig. C.2a. For methods based on point clouds (Choy et al., 2019; Ma et al., 2022; Guo et al., 2021), labeled tree instances are extracted from the full scene and uniformly downsampled to 8,192 points per instance. As for the multi-modal method TSCMDL (Liu et al., 2023a), camera poses are used to spatially associate tree instances with corresponding panoramic images, followed by cropping of image patches based on projected 2D bounding boxes. These image patches are resized to 768×512 when input into TSCMDL.

Fig. C.2b illustrates the input data processing workflow for tree segmentation baselines. Specifically, for the uni-modal methods (Wielgosz et al., 2024; Vu et al., 2022; Sun et al., 2023), point clouds are processed using cylindrical sampling (16m radius), implemented via a greedy algorithm to ensure comprehensive coverage. Additionally, random sampling is introduced to enrich the training dataset. During testing, we utilize cylindrical blocks of the same radius and performed uniform sampling with a fixed step size to guarantee full scene coverage. The multi-modal network LCPS (Zhang et al.,

Table B.1
Abbreviations of tree species.

| Tree species | Abbreviation |
|---|---|
| *Platanus × acerifolia* | PA |
| *Cinnamomum camphora* | CC |
| *Lagerstroemia indica* | LI |
| *Osmanthus fragrans* | OF |
| *Koelreuteria paniculata* | KP |
| *Prunus cerasifera 'atropurpurea'* | PC |
| *Zelkova serrata* | ZS |
| *Cedrus deodara* | CD |
| *Ginkgo biloba* | GB |
| *Acer palmatum var. atropurpureum* | AA |
| *Prunus serrulata* | PS |
| *Ligustrum lucidum* | LL |
| *Malus × micromalus* | MM |
| *Magnolia grandiflora* | MG |
| *Metasequoia glyptostroboides* | MS |
| *Sophora japonica* | SJ |
| *Acer pictum subsp. mono* | AM |
| *Photinia × fraseri* | PF |

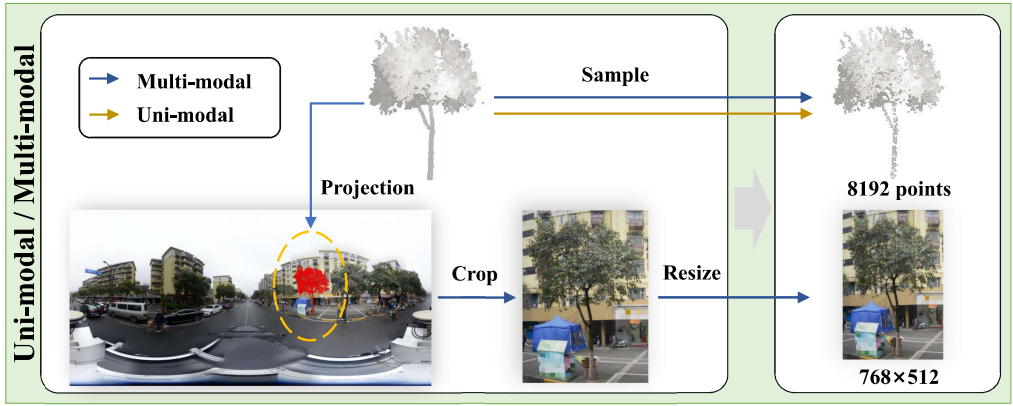

(a) Input data processing workflow for species classification

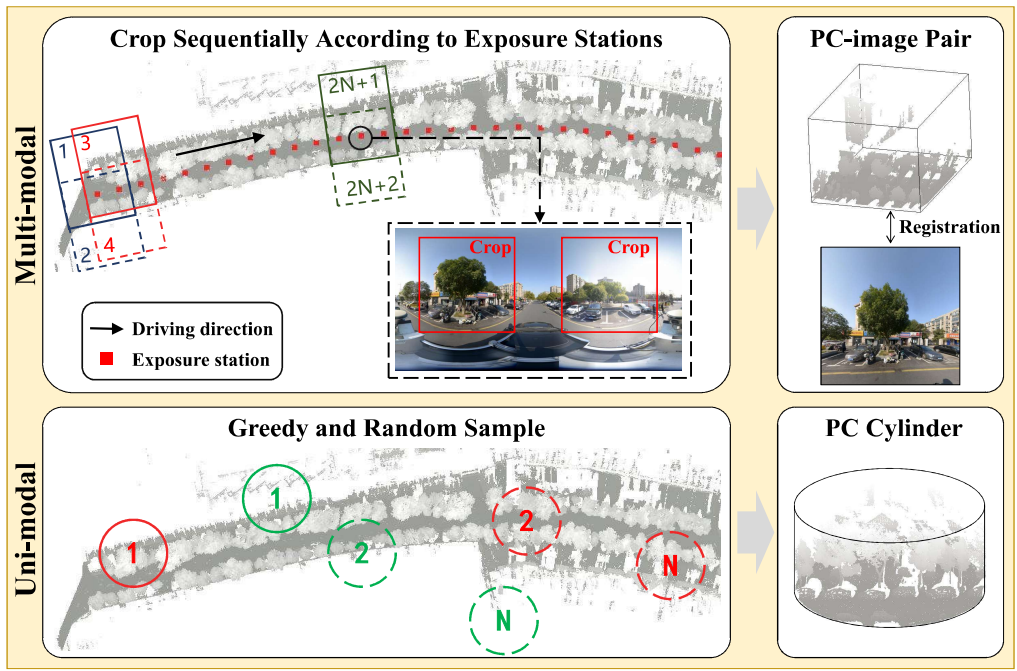

(b) Input data processing workflow for tree segmentation

Fig. C.2. Details of input data processing workflow for tree species classification and individual tree segmentation baselines.

2023) requires specialized preprocessing. We convert raw panoramic images into frame-based images compatible with mainstream image backbones, retaining only a 120° horizontal and vertical FOV to minimize lateral distortion while preserving roadside tree distributions. The corresponding point clouds are cropped using a 30m×40m square instead of the image frustum to avoid scene omissions caused by narrow FOVs, thereby ensuring spatial alignment of multi-modal inputs. During inference, predictions from overlapping blocks are fused using the method proposed by Xiang et al. (2023). Subsequent block merging and upsampling are performed to ensure complete coverage, enabling predictions to be assigned to all points in the original road segments.

All baseline implementations strictly follow the original experimental configurations, with minor adaptations made for compatibility with our dataset.

**Appendix D. Supplementary experimental results**

Table D.2 shows detailed per-species results of tree species classification baselines.

Fig. D.3 - Fig. D.6 show confusion matrix of tree species classification baselines.

Fig. D.7 and Fig. D.8 present the qualitative results of tree segmentation baseline algorithms on WHU-STree-NJ and WHU-STree-SY datasets.

The detailed per-species results of each tree segmentation baseline algorithm are presented in Table D.3 - Table D.6.

Table D.2
Detailed results of the species classification benchmark on WHU-STree-NJ. Per-species IoU (%) are reported.

| Method | MinkNet | PointMLP | PTv2 | TSCMDL |
|---|---|---|---|---|
| PA | 89.92 | 91.65 | **95.73** | 94.11 |
| CC | 79.12 | 81.00 | 86.46 | **87.23** |
| LI | 70.51 | 69.41 | **81.84** | 77.94 |
| OF | 75.49 | 78.87 | **81.37** | 71.96 |
| KP | **64.42** | 44.26 | 55.73 | 50.86 |
| PC | 34.45 | 36.89 | **52.68** | 43.81 |
| ZS | 50.91 | 56.00 | **63.64** | 49.18 |
| CD | 57.32 | **90.43** | 85.83 | 88.43 |
| GB | 61.54 | 60.56 | **66.67** | 60.71 |
| AA | 53.54 | 49.38 | 48.24 | **54.76** |
| PS | 36.89 | **54.35** | 48.15 | 51.25 |
| LL | 1.41 | 19.48 | **46.51** | 45.12 |
| MM | 44.93 | 52.04 | **66.23** | 39.60 |
| MG | 70.00 | 86.21 | 80.65 | **91.38** |
| MS | 61.19 | 95.58 | **95.59** | 84.85 |
| SJ | 0.00 | 9.43 | **24.53** | 22.81 |
| AM | 42.86 | **52.63** | 49.58 | 51.54 |
| PF | 36.92 | 48.15 | **55.77** | 50.94 |
| Others | 17.78 | 25.76 | 32.53 | **32.86** |

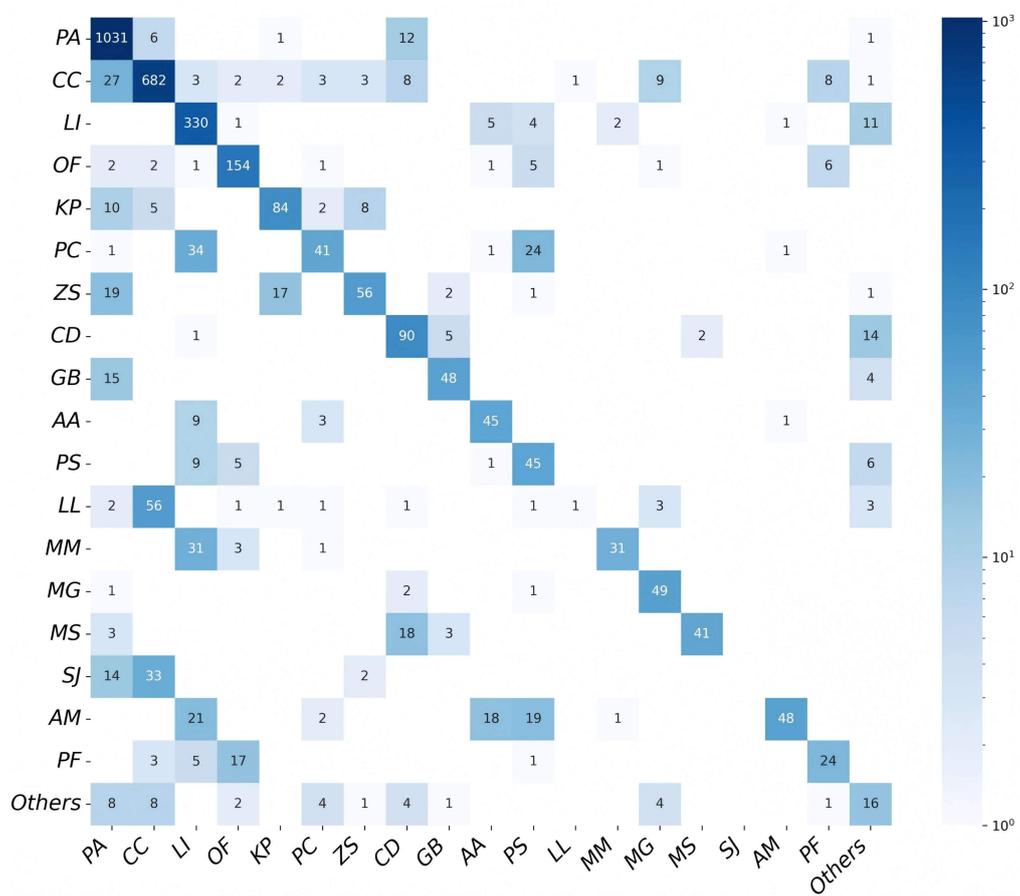

Fig. D.3. Confusion matrix of MinkNet on WHU-STree-NJ dataset (Rows: True species; Columns: Predicted species).

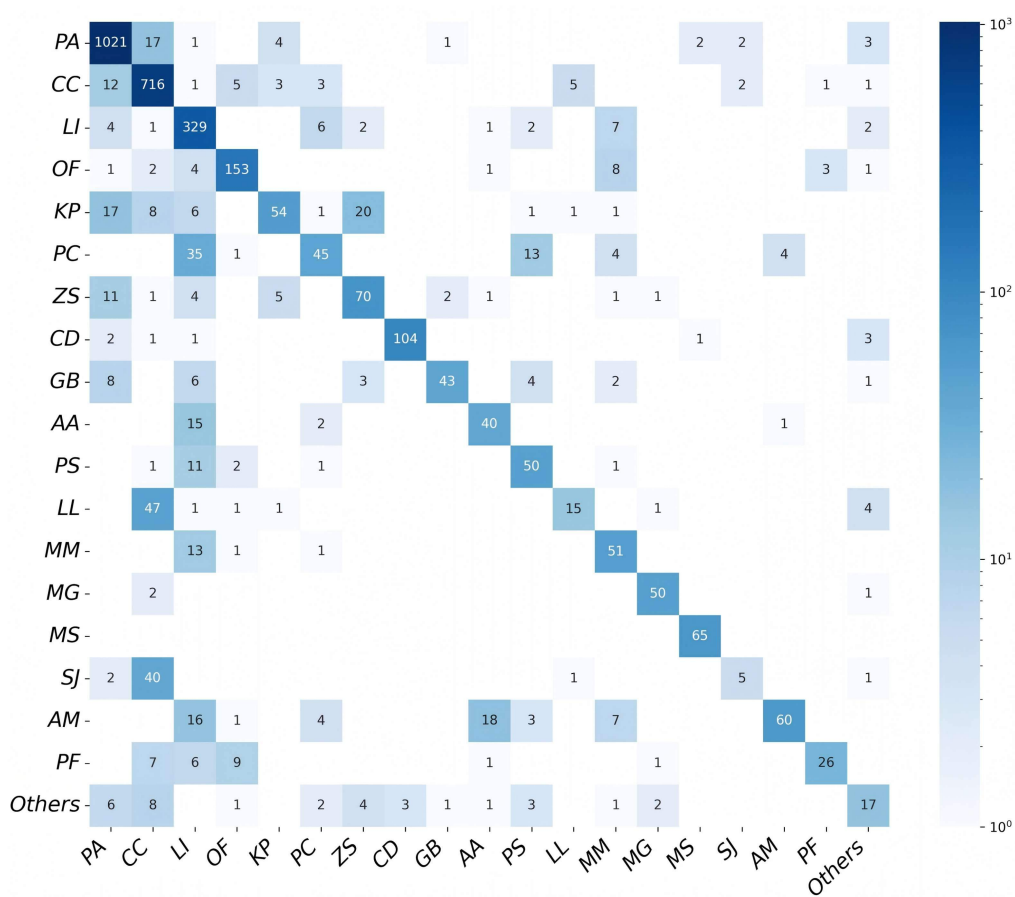

Fig. D.4. Confusion matrix of PointMLP on WHU-STree-NJ dataset (Rows: True species; Columns: Predicted species).

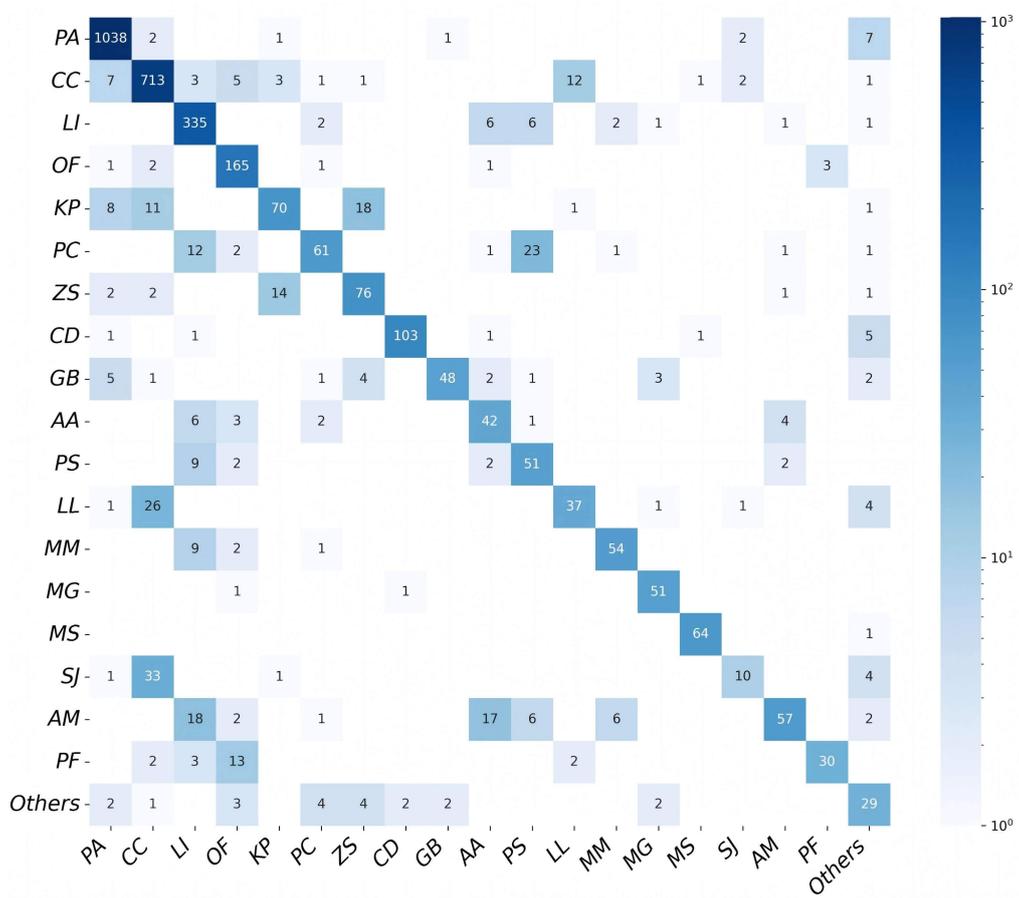

Fig. D.5. Confusion matrix of PTv2 on WHU-STree-NJ dataset (Rows: True species; Columns: Predicted species).

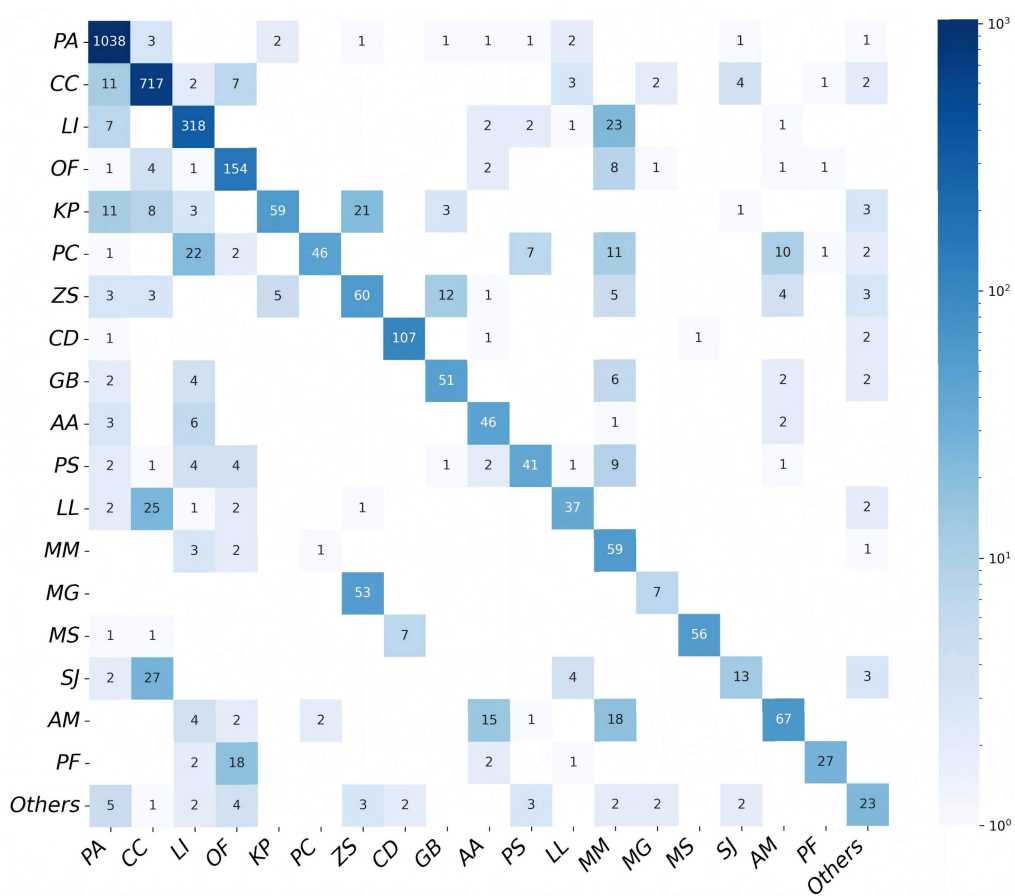

Fig. D.6. Confusion matrix of TSCMDL on WHU-STree-NJ dataset (Rows: True species; Columns: Predicted species).

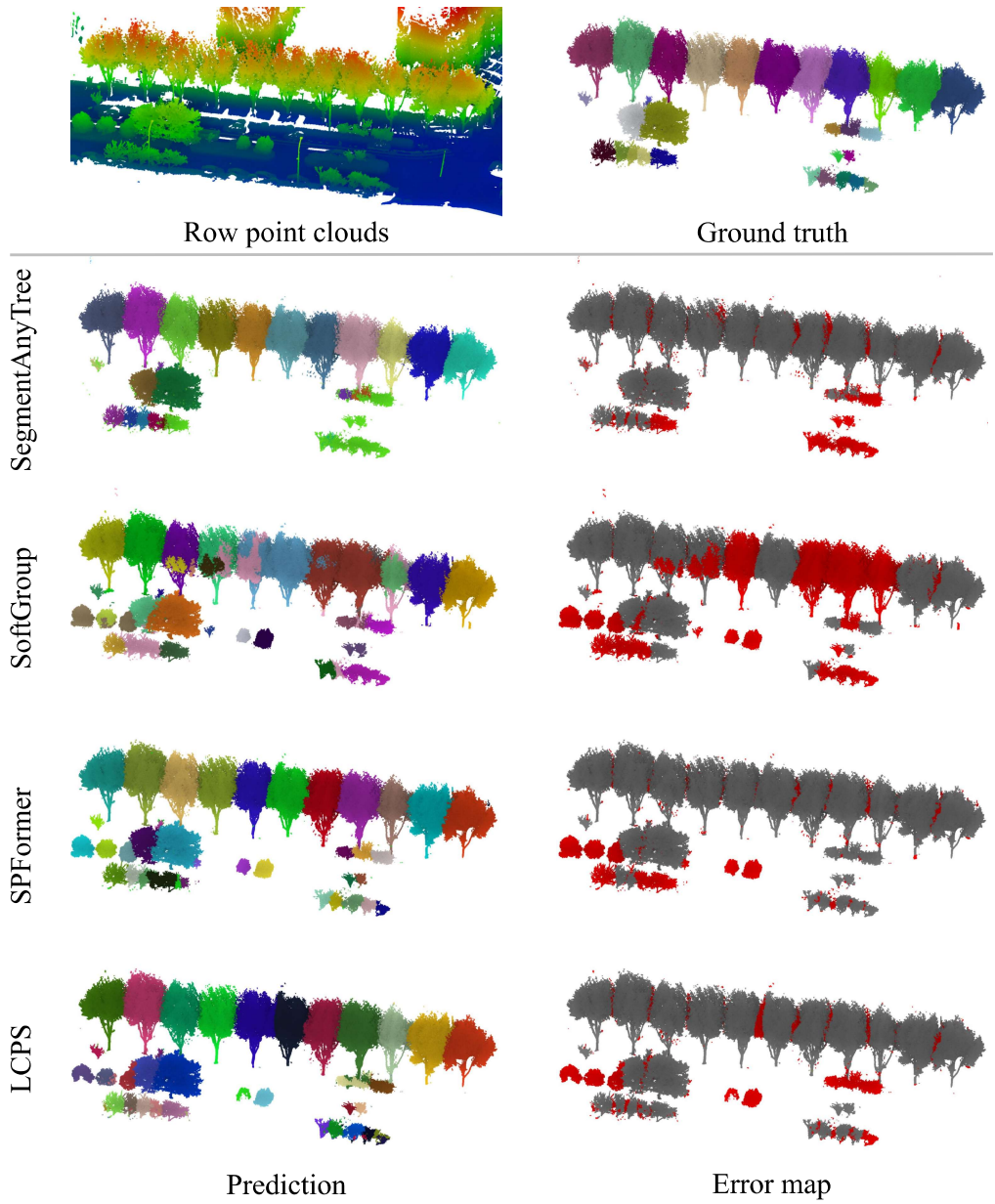

Fig. D.7. Visual results of tree segmentation benchmark on WHU-STree-NJ. Red areas in the error map denote erroneous segments.

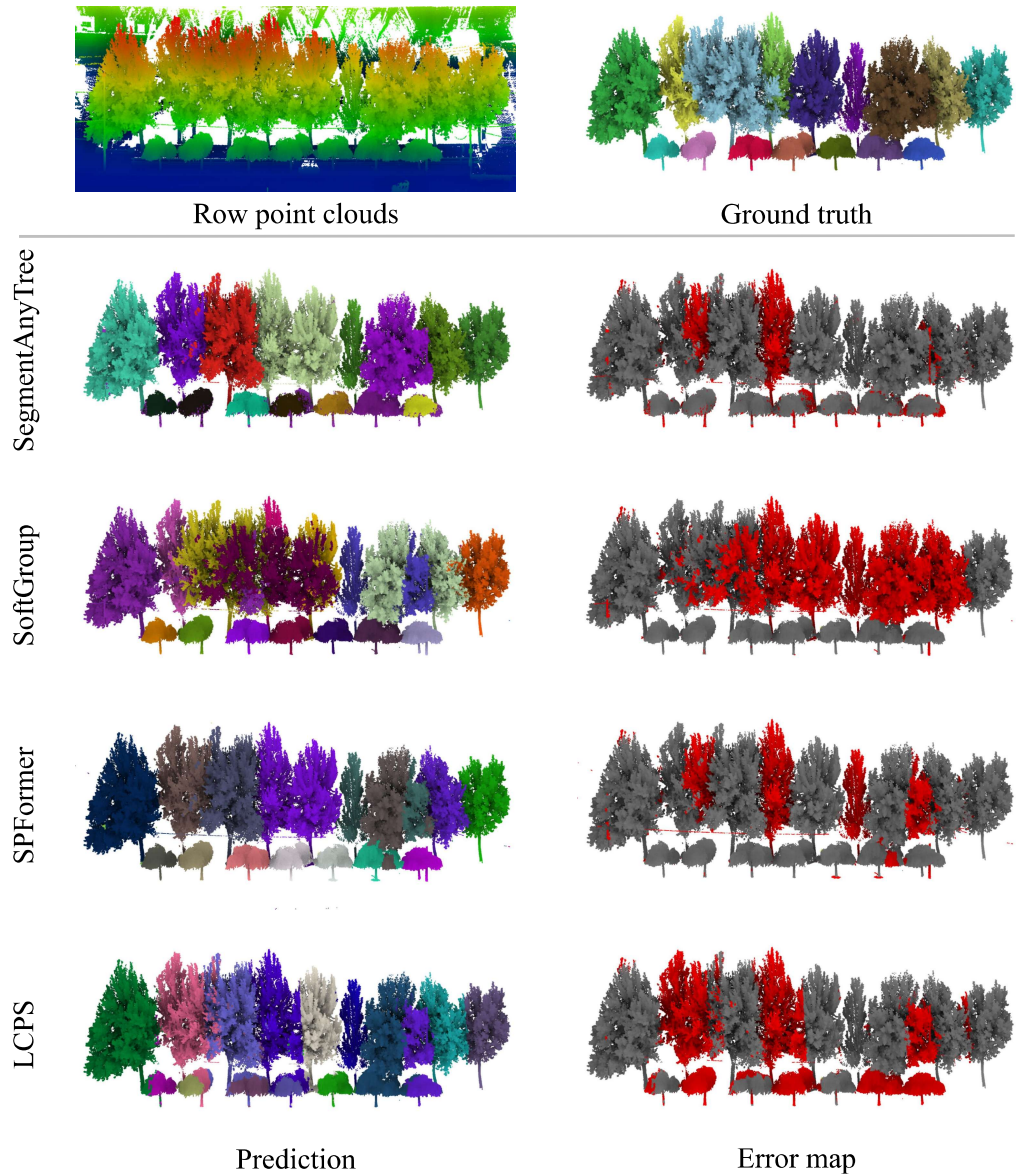

Fig. D.8. Visual results of tree segmentation benchmark on WHU-STree-SY. Red areas in the error map denote erroneous segments.

Table D.3
Detailed results of SegmentAnyTree on WHU-STree-NJ dataset.

|        | Cov↑(%) | WCov↑(%) | Prec↑(%) | Rec↑(%) |
|--------|---------|----------|----------|---------|
| PA     | 60.2    | 71.5     | 64.9     | 64.3    |
| CC     | 84.1    | 86.4     | 86.2     | 90.3    |
| LI     | 84.8    | 84.6     | 51.0     | 90.4    |
| OF     | 52.9    | 53.1     | 60.4     | 53.8    |
| KP     | 72.8    | 80.2     | 85.9     | 74.5    |
| PC     | 39.3    | 55.1     | 35.9     | 39.8    |
| ZS     | 64.9    | 66.5     | 31.8     | 72.9    |
| CD     | 52.2    | 56.7     | 88.5     | 59.0    |
| GB     | 41.8    | 72.4     | 95.7     | 44.0    |
| AA     | 0.0     | 0.0      | 0.0      | 0.0     |
| PS     | 31.9    | 16.2     | 26.6     | 33.3    |
| LL     | 58.3    | 65.4     | 42.1     | 65.6    |
| MM     | 3.3     | 3.1      | 12.5     | 2.3     |
| NG     | 0.0     | 0.0      | 0.0      | 0.0     |
| NS     | 0.0     | 0.0      | 0.0      | 0.0     |
| SJ     | 0.0     | 0.0      | 0.0      | 0.0     |
| AM     | 0.0     | 0.0      | 0.0      | 0.0     |
| PF     | 34.8    | 37.2     | 100.0    | 36.7    |
| Others | 52.5    | 71.3     | 8.9      | 55.3    |

Table D.4
Detailed results of SoftGroup on WHU-STree-NJ dataset.

|        | Cov↑(%) | WCov↑(%) | Prec↑(%) | Rec↑(%) |
|--------|---------|----------|----------|---------|
| PA     | 57.9    | 58.7     | 58.5     | 53.4    |
| CC     | 68.8    | 67.7     | 66.2     | 68.5    |
| LI     | 44.0    | 49.9     | 47.9     | 36.2    |
| OF     | 48.1    | 45.2     | 37.4     | 39.3    |
| KP     | 87.4    | 57.5     | 62.7     | 88.9    |
| PC     | 68.6    | 73.2     | 52.5     | 69.7    |
| ZS     | 67.1    | 71.1     | 56.8     | 74.0    |
| CD     | 56.0    | 60.5     | 44.5     | 58.0    |
| GB     | 47.1    | 69.3     | 88.9     | 48.5    |
| AA     | 0.0     | 0.0      | 0.0      | 0.0     |
| PS     | 0.0     | 0.0      | 0.0      | 0.0     |
| LL     | 22.2    | 29.9     | 84.6     | 23.9    |
| MM     | 0.0     | 0.0      | 0.0      | 0.0     |
| NG     | 0.0     | 0.0      | 0.0      | 0.0     |
| NS     | 5.7     | 3.0      | 80.0     | 6.3     |
| SJ     | 0.0     | 0.0      | 0.0      | 0.0     |
| AM     | 0.0     | 0.0      | 0.0      | 0.0     |
| PF     | 0.0     | 0.0      | 0.0      | 0.0     |
| Others | 23.4    | 15.5     | 4.8      | 24.5    |

Table D.5
Detailed results of SPFormer on WHU-STree-NJ dataset.

|        | Cov↑(%) | WCov↑(%) | Prec↑(%) | Rec↑(%) |
|--------|---------|----------|----------|---------|
| PA     | 81.4    | 83.0     | 73.0     | 92.0    |
| CC     | 88.1    | 88.4     | 69.1     | 95.1    |
| LI     | 89.3    | 90.2     | 67.3     | 95.5    |
| OF     | 83.5    | 80.4     | 44.3     | 86.6    |
| KP     | 87.4    | 78.9     | 75.4     | 88.7    |
| PC     | 9.2     | 11.9     | 4.8      | 11.1    |
| ZS     | 68.3    | 76.6     | 68.0     | 74.7    |
| CD     | 80.8    | 85.9     | 78.9     | 90.2    |
| GB     | 78.1    | 82.3     | 88.2     | 78.9    |
| AA     | 39.6    | 34.4     | 49.0     | 42.9    |
| PS     | 81.7    | 87.5     | 46.1     | 89.4    |
| LL     | 16.5    | 23.7     | 38.5     | 17.5    |
| MM     | 68.1    | 74.0     | 30.8     | 72.7    |
| NG     | 5.6     | 8.0      | 33.3     | 5.9     |
| NS     | 80.5    | 83.7     | 92.1     | 90.6    |
| SJ     | 0.0     | 0.0      | 0.0      | 0.0     |
| AM     | 47.6    | 62.1     | 76.4     | 51.9    |
| PF     | 89.5    | 90.4     | 61.7     | 96.7    |
| Others | 32.5    | 31.9     | 6.3      | 33.3    |

Table D.6
Detailed results of LCPS on WHU-STree-NJ dataset.

|  | Cov↑(%) | WCov↑(%) | Prec↑(%) | Rec↑(%) |
|---|---|---|---|---|
| PA | 76.1 | 80.7 | 78.9 | 87.2 |
| CC | 80.6 | 83.8 | 71.6 | 86.7 |
| LI | 20.0 | 13.8 | 68.5 | 19.2 |
| OF | 8.5 | 13.8 | 19.4 | 7.4 |
| KP | 91.3 | 90.3 | 86.7 | 92.9 |
| PC | 65.1 | 72.9 | 36.3 | 67.3 |
| ZS | 64.9 | 41.6 | 73.8 | 68.1 |
| CD | 0.0 | 0.0 | 0.0 | 0.0 |
| GB | 91.4 | 92.4 | 94.7 | 94.7 |
| AA | 37.1 | 34.3 | 17.9 | 39.3 |
| PS | 30.0 | 31.7 | 28.4 | 31.8 |
| LL | 24.3 | 23.4 | 42.9 | 33.3 |
| MM | 0.0 | 0.0 | 0.0 | 0.0 |
| NG | 0.0 | 0.0 | 0.0 | 0.0 |
| NS | 0.0 | 0.0 | 0.0 | 0.0 |
| SJ | 0.0 | 0.0 | 0.0 | 0.0 |
| AM | 14.3 | 21.3 | 71.4 | 14.2 |
| PF | 34.9 | 36.1 | 100.0 | 43.3 |
| Others | 32.7 | 29.5 | 6.2 | 36.7 |